\documentclass{article}

\usepackage[preprint]{neurips_2021}
\usepackage{graphicx}
\usepackage[linesnumbered]{algorithm2e}
\usepackage{todonotes}
\usepackage{url}
\usepackage{booktabs}
\usepackage{amsfonts}
\usepackage{caption}
\usepackage{subcaption}
\usepackage{mathbbol}
\usepackage{multirow}
\usepackage{graphicx, array, blindtext}
\usepackage{amsmath}
\usepackage{sidecap}
\usepackage{sidecap}
\usepackage{caption,subcaption}
\usepackage{pbox}
\usepackage{arydshln}
\newtheorem{observation}{Observation}
\newtheorem{proposition}{Proposition}

\DeclareMathOperator*{\argmax}{arg\,max}
\def\bs#1{\boldsymbol{#1}}
\def\set#1{\bs{#1}}

\newcommand{\numnodes}{N}
\newcommand{\unode}{i}
\newcommand{\vnode}{j}

\newcommand{\xinstance}{\set{x}}

\newcommand{\linstance}{\set{z}}
\newcommand{\ldim}{t}
\newcommand{\lthGraph}{l}

\newcommand{\ndim}{d} 

\newcommand{\decode}{\textsc{DECODE}}

\newcommand{\GraphNum}{S}
\newcommand{\nodeNum}{N}
\newcommand{\nodem}{i}
\newcommand{\noden}{j}
\newcommand{\G}{G}
\newcommand{\A}{\set{A}} 
\newcommand{\D}{D} 
\newcommand{\X}{\set{X}} 
\newcommand{\Z}{\set{Z}}
 
\newcommand{\prob}[1]{\Tilde{#1}}

\newcommand{\V}{V} 
\newcommand{\dhk}{\textbf{d}}
\newcommand{\baseline}{FC}

\newcommand{\eparameters}{\varphi} 
\newcommand{\dparameters}{\theta} 

\newcommand{\kernel}{k}

\newcommand{\mmd}{\mathit{D}^{2}}
\newcommand{\lmmd}{L_{\textsc{kelbo}}}
\newcommand{\lelbo}{L_{\textsc{elbo}}}
\newcommand{\kthFeature}{u}
\newcommand{\featureNUM}{m} 
\newcommand{\kweight}{\lambda}

\newcommand{\elbo}{\textsc{ELBO}~}
\newcommand{\featureFunction}{\phi}

\usepackage[utf8]{inputenc} 
\usepackage[T1]{fontenc}    
\usepackage{hyperref}       
\usepackage{url}            
\usepackage{booktabs}       
\usepackage{amsfonts}       
\usepackage{nicefrac}       
\usepackage{microtype}      
\PassOptionsToPackage{numbers, compress}{natbib}

\title{Generating the Graph Gestalt: Kernel-Regularized Graph Representation Learning}
\author{ Kiarash Zahirnia, Ankita Sakhuja, Oliver Schulte, Parmis Nadaf, Ke Li, Xia Hu }

\begin{document}

\maketitle

\begin{abstract}\label{abstract}
Recent work on graph generative models has made remarkable progress towards generating increasingly realistic graphs, as measured by global graph features such as degree distribution, density, and clustering coefficients. 
Deep generative models have also made significant advances through better modelling of the local correlations in the graph topology, which have been very useful for predicting unobserved graph components, such as the existence of a link or the class of a node, from nearby observed graph components. 
A complete scientific understanding of graph data should address both global and local structure. 
In this paper, we propose a joint model for both as complementary 
objectives in a graph VAE framework. Global structure is captured by incorporating graph kernels in a probabilistic model whose loss function is closely related to the maximum mean discrepancy (MMD) between the global structures of the reconstructed and the input graphs. 
The \elbo objective derived from the model regularizes a standard local link reconstruction term with an MMD term. 
Our experiments demonstrate a significant improvement in the realism of the generated graph structures, typically by 1-2 orders of magnitude of graph structure metrics, compared to leading graph VAE and GAN models. Local link reconstruction  improves as well in many cases. 
%
\end{abstract}

\section{Introduction}\label{Introduction}
Many important datasets contain relational information about entities and their links that can be represented as a graph. 
Deep generative learning on graphs has become a popular research topic~\cite{hamilton2020graph}, with many applications including molecule design~\cite{DBLP:journals/jmlr/SamantaDJGCGG20}, matrix completion for inference, and recommendation~\cite{do2019matrix}. Computational methods for analysing graph data have developed two distinct traditions~\cite{bib:chakrabarti2012graph}, that emphasize (i) local correlations in the graph topology~\cite{kipf2016variational,arga,graphite} (ii) global aspects of the graph structure. The paper develops a new approach to jointly modelling both local correlations and global graph structure in a deep generative  framework. 

Typical graph-based local predictions are predicting the existence of a link between two nodes from node attributes and other links, and predicting a class attribute of a node~\cite{hamilton2020graph}. For example, in a social network, a typical task is to recommend potential new friends based on existing friendship links. An example of node classification is to predict the political affiliation of a user given their friendships and affiliations of their friends. Deep graph representation learning has achieved excellent performance on local prediction tasks by computing embeddings, or latent features, of nodes or edges~\cite{kipf2016variational,arga,graphite}. 

\begin{figure}[h]
     \centering
     \begin{subfigure}[b]{.3\textwidth}
         \centering
         \includegraphics[width=\textwidth]{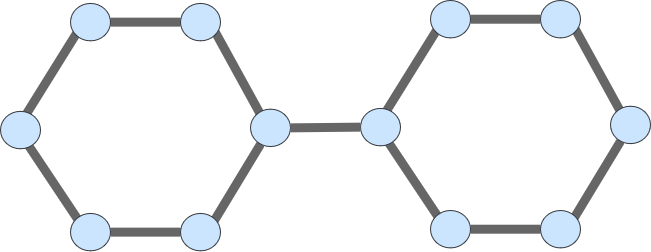}
                  \caption{Original Graph}
         \label{fig:original}
     \end{subfigure}
     \hfill
     \begin{subfigure}[b]{0.3\textwidth}
         \centering
         \includegraphics[width=\textwidth]{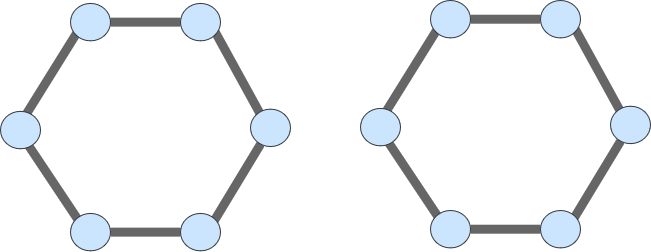}
                  \caption{Generated Graph 1}
         \label{fig:sample1}
     \end{subfigure}
     \hfill
     \begin{subfigure}[b]{0.3\textwidth}
         \centering
         \includegraphics[width=\textwidth]{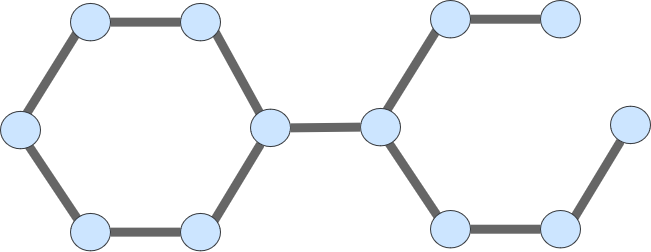}
                  \caption{Generated Graph 2}
         \label{fig:sample2}
     \end{subfigure}
     \caption{The difference between local and global structure. The two right graphs~(\ref{fig:sample1}),~(\ref{fig:sample2}) score the same in terms of number of edges reconstructed from the left graph~(\ref{fig:original}). 
    However the middle graph~(\ref{fig:sample1}), is structurally more similar to~(\ref{fig:original}), containing the same number of connected components.}
        \label{fig:crossentropyExample}
\end{figure}

Structural generative graph models, often drawing on network analysis~\cite{bib:chakrabarti2012graph}, aim to capture global graph features~\cite{you2018graphrnn,li2018learning,DBLP:journals/jmlr/SamantaDJGCGG20}, such as degree distribution, density, and clustering coefficients. We refer to these properties collectively as the \emph{graph gestalt}; gestalt is a term from psychology that denotes the overall organization of a set of objects that is more than just the set of its parts.  Figure~\ref{fig:crossentropyExample} illustrates the concept.
For another example, in a grid graph, small cliques of edges are not distinctive in themselves, but collectively form a highly regular pattern. 
 \citet{you2018graphrnn} have shown empirically that local representation learning provides a poor model of graph gestalt: graphs generated from local embeddings have an unrealistic global structure. The reason for this is that {\em link reconstruction is inherently local}: the objective function is based on element-wise local comparisons between the observed and the predicted  adjacency matrix. In a graph VAE (GVAE), the local link reconstruction probability is the product of link probabilities that are conditionally independent given node embeddings. While conditional independence supports easy and stable optimization, it measures only local data fit and not the global graph properties.
 Moreover, the link reconstruction probability weights all edges equally, but as Figure~\ref{fig:crossentropyExample} illustrates, not all edges in the graphs are equally important~\cite{faisal2015edge}. A complete model of graph data should capture both global and local properties, which can reinforce each other in a joint model to improve both. 
 Modelling global structure supports inductive graph learning~\cite{you2018graphrnn} when the graph gestalt is a stable high-level pattern that generalizes from subgraphs to a larger graph. 
 
{\em Approach.} 
We utilize a joint probabilistic model over both individual graph adjacencies and global graph features. Graph features are represented by user-specified graph kernels. The \elbo objective of the model regularizes a standard local adjacency reconstruction term with a maximum mean distance (MMD) term derived from the graph kernels; we refer to it as the {\em kernel-regularized} \elbo. Kernel regularization connects graph representation learning with  the rich body of work on graph kernels. We show that the kernel-regularized \elbo has two complementary interpretations: i) As an adjacency reconstruction objective with a graph moment matching constraint. ii) As a joint reconstruction objective over both adjacencies and graph features.

{\em Evaluation.} Our evaluation compares training GVAEs with and without kernel regularization. We keep architecture, datasets, and evaluation metrics the same and change only the objective function. Our comparison methods include three GVAEs, two state-of-the-art and one benchmark vanilla GVAE. We also benchmark against a SOTA graph GAN and auto-regressive  architectures designed for generating realistic graphs ~\cite{you2018graphrnn,DBLP:conf/nips/LiaoLSWHDUZ19}. 
The input graph kernels are based on node degree and transition probabilities. For GVAEs, kernel-regularization improves the reconstruction of graph structure metrics by orders of magnitude. Compared to auto-regressive  architectures for generating realistic graphs, kernel-regularization generates graph structure of competitive quality, and scales much better to larger graphs. 
We also provide evidence that matching input graph structure often has the effect of a co-training objective for local graph modelling, improving local link reconstruction, and increasing test predictive accuracy for node classification. 

{\em Contributions.} Our contributions can be summarized as follows. 
\begin{itemize}
\item A new joint probabilistic model over both individual graph adjacencies and global graph features.
\item Deriving a kernel-regularized \elbo as a new objective function for training graph variational auto-encoders.
\item Differentiable extensions of previous graph kernels to  so they can be applied to reconstructed probabilistic adjacency matrices.
\end{itemize}
 \section{Related Work}
 
 We briefly discuss the relationship to the most relevant work.
 
 {\em Graph Neural Networks.} This paper investigates several deep generative graph models, including those based on GANs, VAEs, and auto-regressive models. Overviews can be found in ~\cite{zhou2020graph,hamilton2020graph}. Graph GANs~\cite{de2018molgan} and VAEs ~\cite{kipf2016variational,graphite,osbm_gnn} use scalable {\em edge-parallel} generation based on conditional independence of edges. On the other hand, auto-regressive architectures~\cite{DBLP:conf/nips/LiaoLSWHDUZ19,you2018graphrnn} generate edges sequentially, which allows them to capture complex dependencies between a new edge and edges already generated. They typically require a canonical node ordering for generating edges in order. Our contribution is not a new model but a new objective function for training generative graph models. While our exposition focuses on kernel-regularization for graph VAEs, in principle it could be used with other architectures as well. 
 
 
 {\em Graph Moment Matching.} As we show in Section~\ref{sec:interpret}, kernel-regularization can be interpreted as a graph moment matching condition, that rewards a model for matching observed graph statistics (e.g. node distribution). Graph moment matching has been previously studied in network statistics; see~\cite[Sec.7]{orbanz2014bayesian}. This work has focused on theoretical properties of graph moment estimators with increasing graph size. Our goal is to describe and evaluate a practical computational architecture within a deep encoder-decoder framework. 
 
 {\em Graph Sparsity.} 
 A well-known issue with edge-parallel models is that they generate unrealistically dense graphs~\cite[Sec.7]{orbanz2014bayesian}, which has motivated graph moment matching.  \citet{kipf2016variational} suggest increasing the weight of existing edges to reduce density, but this reduces overall graph reconstruction quality~\cite{qiao2020novel}. Our experiments show that kernel regularization offers a principled approach for reducing edge-parallel density.
 
 {\em Constrained \elbo Maximization.} \citet{DBLP:conf/nips/MaCX18}
describe general constrained \elbo maximization using quadratic constraints comparable to our MMD term. They require the constraint to hold for almost all  latent 
variable instances $\linstance$ with measure 1 in the {\em prior} distribution $p(\Z)$, whereas our formulation requires the constraint to hold for almost all latent variables instances in the {\em posterior} distribution $q_{\eparameters}(\Z|\X,\A)$. This difference arises because the goal of \citeauthor{DBLP:conf/nips/MaCX18} is to capture a prior constraint valid for all graphs in a domain, whereas our goal is to reconstruct a specific input graph.
 
 {\em Graph Kernels.} We define differentiable kernels for two graph aspects: the node degree distribution and transition probabilities from random walks. For node degrees, we adapt the vector label histogram (VLH) graph kernel \citet{nikolentzos2019graph}, which guarantees permutation invariance (i.e., the degree histogram does not depend on the ordering of the nodes). The main difference is that VLH was originally defined for discrete node labels whereas we apply it with degrees as continuous node features. Utilizing multi-step transition probability matrices 
 as a basis for a graph kernel was explored by \citet{DBLP:conf/nips/0007WX0N18}. The difference is that we use a simpler kernel to compare transition probability matrices but utilize more information, especially the off-diagonal entries.

\section{Data Model and \elbo Objective}\label{theory}
An attributed graph is a pair $\G=(V,E)$ comprising a finite set of nodes and edges where each node  is assigned an $\ndim$-dimensional attribute $\xinstance_{\unode}$ with $\ndim>0$. An attributed graph can be represented by an $\numnodes\times \numnodes$ adjacency matrix $\A$ with $\{0,1\}$ Boolean entries, together with an $\numnodes \times \ndim$ node feature matrix $\X$. 
Following \citet{DBLP:conf/nips/MaCX18}, we view the observed adjacency matrix as a sample from an underlying probabilistic adjacency matrix $\prob{\A}$ with $\prob{\A}_{\nodem,\noden} \in [0,1]$. The sampling distribution for independent edges  is given by
\begin{equation} \label{eq:sample-edges}
    P(\A|\prob{\A}) = 
    \prod_{\nodem=1}^{\numnodes} \prod_{\noden=1}^{\numnodes} \prob{\A}_{\nodem,\noden}^{\A_{\nodem,\noden}} (1-\prob{\A}_{\nodem,\noden})^{1-\A_{\nodem,\noden}}.
\end{equation}
A graph kernel $k(\G,\G')$ measures the similarity of two (attributed) graphs.  For a given graph kernel $\kernel$ and two graph sets 
$\set{G}=\{\G_{\lthGraph}\}_{\lthGraph=1}^{\GraphNum}$ and $\set{G'}=\{\G'_{\lthGraph})\}_{\lthGraph=1}^{M}$ the \textbf{maximum mean distance} is defined as $\mathit{MMD}^2_{\kernel}$ as usual between two sets (see~\cite{you2018graphrnn}).
%
For the special case of comparing a single graph to another ($\GraphNum=M=1$), we write 
\[\mmd_{\kernel}(\G,\G') \equiv \mathit{MMD}^2_{\kernel}(\{\G\},\{\G'\}) = \kernel(\G_\lthGraph,\G_{\lthGraph}) + \kernel(\G'_\lthGraph,\G'_{\lthGraph}) - 2\kernel(\G_\lthGraph,\G'_{\lthGraph}).\] We also write $\kernel(\A,\A')$ and $\mmd_{\kernel}(\A,\A')$  if we wish to emphasize that a graph kernel and MMD can be computed as a function of the adjacencies only.





\subsection{Kernel-Regularized \elbo Objective}
\begin{figure}[h]
\centering
\includegraphics[width=0.30\textwidth]{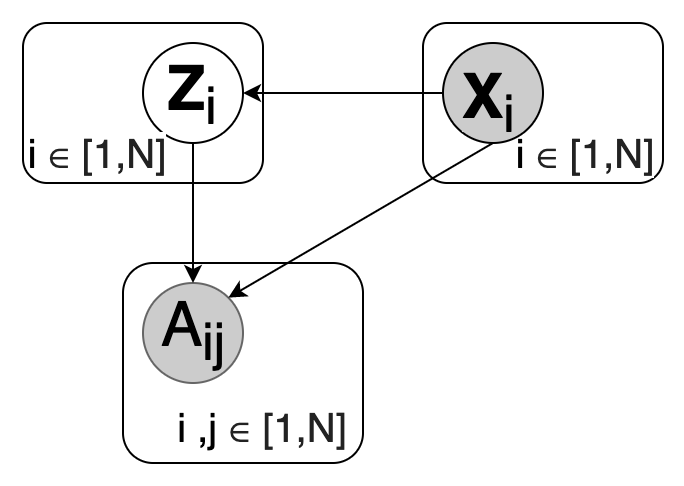}
\caption{Latent variable model for VAE. Observed evidence
variables in gray.}
\label{fig:latent_variable}
\end{figure}

Figure~\ref{fig:latent_variable} shows the standard generative graphical model for GVAEs~\cite{kipf2016variational,graphite}. 
We formulate a new variational objective function in a generic encoder-decoder framework with the following properties. 
\begin{itemize}
    \item The encoder deterministically maps an input graph representation $(\A,\X)$ to a posterior latent distribution $q_{\eparameters}(\Z|\X,\A)$ where $\Z_{\numnodes \times \ldim}$ specifies a latent $\ldim$-dimensional feature vector for each node $\unode$. The encoder parameters to be optimized are denoted $\eparameters$.
    \item The decoder deterministically maps a latent representation $\linstance$ to a probabilistic graph $\prob{\A}_{\linstance}$. The decoder parameters to be optimized are denoted $\dparameters$. 
    \item A list of differentiable graph kernels $\kernel_{1},\ldots,\kernel_{\featureNUM}$ where $\kernel_{\kthFeature}(\A,\A')$ measures the similarity of two graphs in terms of their adjacency matrices. 
    Each kernel $\kernel_{\kthFeature}$ can be applied to general weighted adjacency matrices, including probabilistic adjacencies~\cite{vishwanathan2010graph}.
\end{itemize}
Combining the mapping $\prob{\A}_{\linstance}$ with the sampling distribution Eq.~\eqref{eq:sample-edges} defines the {\em (local) link reconstruction probability} denoted $p_{\dparameters}(\A|\prob{\A}_{\linstance})$. 
The notation $\lelbo$ denotes the standard \elbo objective for a graph VAE~\cite{kipf2016variational} and 
$\lmmd$ our new 
\textbf{kernel \elbo} objective that adds an MMD term to the link reconstruction probability. 
\begin{small}
\begin{align}
     \lmmd(\eparameters,\dparameters,\set{\kweight}) & \equiv & E_{\linstance \sim q_{\eparameters}(\Z|\X,\A)}\big[\ln p_{\dparameters}(\A|\prob{\A}_{\linstance}) - \sum_{\kthFeature=1}^{\featureNUM} \set{\kweight}_{\kthFeature} \mmd_{\kernel_{\kthFeature}}(\A,\prob{\A}_{\linstance})\big] -KL\big(q_{\eparameters}(\Z|\X,\A)||p(\Z)\big) \label{eq:elbo-augment} 
\end{align}
\end{small}
where $p(\Z)$ is a prior distribution over the latent space. We treat the kernel weights $\set{\kweight}$ as a hyperparameter and  maximize $\lmmd$ with respect to the encoder-decoder parameters $\eparameters,\dparameters$.
The next subsection shows that the kernel objective 
is a lower bound on the graph log-likelihood for two complementary generative models. 

\subsection{Interpretation of the Kernel \elbo} \label{sec:interpret}
For notational simplicity, we focus on the case of one graph kernel $\kernel$. By Mercer's theorem, a kernel defines implicit feature map $\featureFunction$ such that $\kernel(\A,\A') = \featureFunction(\A)\bullet\featureFunction(\A')$. 
We can think of $\featureFunction$ as extracting a structural feature vector of a graph. The $\mmd_{\kernel}$ term represents the Euclidean distance between the input graph feature vector and the reconstruction feature vector: $ \mmd_{\kernel}(\A,\A') = ||\featureFunction(\A) - \featureFunction(\A')||^{2}.$

The $\mmd_{\kernel}$ term therefore encourages the model to match the observed structural features. We next show that objective~\eqref{eq:elbo-augment} is a lower bound on the data log-likelihood under two probabilistic models: 1) adding a {\em feature matching constraint} to the standard graph VAE \elbo. 2) A {\em data augmentation model} that defines a {\em joint reconstruction probability} for both observed adjacencies and graph features.
\paragraph{Feature Matching Constraint}
We consider maximizing the standard graph VAE \elbo~\cite{kipf2016variational} with a feature matching constraint: 
Taking the expectation with respect to the posterior, we seek to maximize the reconstruction log-probability $p(\A|\prob{\A}_{\linstance})$ subject to 0 expected MMD:
\begin{small}
\begin{equation} \label{eq:zero-distance}
   \argmax_{\dparameters,\eparameters} E_{\linstance \sim q_{\eparameters}(\Z|\X,\A)}\big[\ln p_{\dparameters}(\A|\prob{\A}_{\linstance})]  -KL\big(q_{\eparameters}(\Z|\X,\A)||p(\Z)\big) \mbox{ s.t. } 0 = E_{\linstance \sim q_{\eparameters}(\Z|\X,\A)}\big[\mmd_{\kernel_{\kthFeature}}(\A,\prob{\A}_{\linstance})] 
\end{equation}
\end{small}
The constraint forces the expected reconstructed graph feature vector to match the input graph features. 
\begin{observation}
The Lagrangian of the constrained optimization problem~\eqref{eq:zero-distance} is equivalent to the kernel $\elbo$~\eqref{eq:elbo-augment}. 
%
\end{observation}
{\em Proof.} The Lagrangian function is \[E_{\linstance \sim q(\Z|\X,\A)}\big[\ln p(\A|\prob{\A}_{\linstance})]  -KL\big(q_{\eparameters}(\Z|\X,\A)||p(\Z)\big) -  \set{\kweight} E_{\linstance \sim q(\Z|\X,\A)}\big[\mmd_{\kernel_{\kthFeature}}(\A,\prob{\A}_{\linstance})].\] Since both expectations are over the same posterior, we can combine them into a single expectation, which yields to Equation~\eqref{eq:elbo-augment}. 
\paragraph{Data Augmentation} We can view the graph features $\featureFunction(\A)$ as part of the observed data, in addition to $\A$ and $\X$, and adopt a product joint reconstruction model:
\begin{align} \label{eq:prod-model}
    p(\A|\linstance)  \propto p(\A|\prob{\A}_{\linstance}) \cdot p(\set{\featureFunction}(\A)|\linstance))
\end{align}
Suppose that the implicit kernel feature space has finite dimension $l$. Then as usual with VAEs, we use an isotropic $l$-dimensional Gaussian reconstruction model centered on the deterministic input reconstruction:
\begin{equation} \label{eq:gaussian-stats}
    p(\featureFunction(\A)|\linstance) = N(\featureFunction(\A)|\featureFunction(\prob{\A}_{\linstance}),\frac{1}{2 \kweight} I) \mbox{ with variance } \kweight \geq 0
\end{equation}
The kernel \elbo is a lower bound on the graph log-likelihood in the joint reconstruction model, up to constant terms independent of the VAE parameters.
\begin{proposition} \label{prop:augmentation}
Assume that the reconstruction probability $ p(\A|\linstance)$ is defined by Equations~\eqref{eq:prod-model} and~\eqref{eq:gaussian-stats}. Then 
\begin{equation}
    \ln{P(\A)}  = \ln{\int_{\linstance} p(\A|\linstance) p(\linstance) d\linstance} \notag  \geq \lmmd(\eparameters,\dparameters) + \frac{l}{2} \ln{2 \kweight} - \frac{l}{2} \ln{2 \pi}. 
    \label{eq:joint-elbo}
\end{equation}
\end{proposition}
The lower bound~\eqref{eq:joint-elbo} differs from the kernel \elbo~\eqref{eq:elbo-augment} only by a term that is independent of the model parameters. The proof is in the appendix. 
\section{Experimental Design: Comparison Methods} 
Our comparison methods include i) GVAE ii) a SOTA Graph GAN method iii) auto-regressive graph generation methods. 

\subsection{Input Graph Kernels} \label{sec:input-kernels}
The kernel \elbo~\eqref{eq:elbo-augment} requires a user-specified input kernel. In our experiments, we utilize simple kernels based on (soft) adjacency matrices that are differentiable with respect to the matrix entries, and easy to interpret: the degree distribution and the $s$-step transition probability matrices, for $s=1,\ldots 5$. 
\paragraph{Degree Distribution Kernel.} For a weighted undirected graph $\prob{\A}$, the \textbf{degree} of node $v_\unode$ is defined as $d(v_\unode) = \sum_{\vnode} \prob{\A}_{\unode\vnode}$. We adapt the vector label histogram  (VLH)  \cite{nikolentzos2019graph}. To form a degree histogram of graph $\prob{\A}$, we bin the soft degrees. Binning creates a histogram comparable to the (hard) degree histogram.  Adopted VLH is based on a soft assignment of points to bins given bin centers and widths. We choose the bin centers to be the possible (hard) node degrees $b=0,\ldots,\numnodes$. All widths are uniformly set to 0.1 (based on experimentation). The soft assignment is then given by $a(v_{\unode},b)=\max \{0, 1 - 0.1 \cdot |d(v_\unode) - b| \} $. Thus the membership of a node in a bin ranges from 0 to 1 and decreases with the difference between the node's expected degree and the bin center. The VLH assigns to each bin the sum of nodes membership in the bin. 
\begin{eqnarray}
   \dhk_{\prob{\A}}(b) \equiv \sum_{\unode = 1}^{\numnodes} a(v_{\unode},b) \label{eq:degree}
   \\
   \kernel_h(\prob{\A}, \prob{\A}') = \dhk_{\prob{A}} \bullet \dhk_{\prob{A}'} \notag
\end{eqnarray}
     In a naive implementation, the computational cost of finding the $\dhk_{\prob{\A}}(b)$ vector
     is $O(n^2)\;$. This can be improved using speed-up techniques for nearest neighbour methods.  
     
     \paragraph{S-Step transition probability kernel.} The $s$-step transition probability matrix of adjacency matrix $\prob{\A}$, is a $\nodeNum \times \nodeNum$ matrix such that ${P^s(\prob{\A})}_{\nodem,\noden}$ is the probability of a transition from node 
     $\nodem$ to node $\noden$ in a random walk of $s$ steps, where the probability of each transition is the normalized link probability. The transition matrix $P^s(\prob{\A})$ can be computed by 
     \begin{equation*}
         P^s(\prob{\A}) = (D(\prob{\A})^{-1}\prob{\A})^{s}
     \end{equation*} 
     where the degree matrix $\D(\prob{\A})$ is a diagonal matrix whose entries $\D(\prob{\A})_{\unode \unode}\equiv d(\unode)$ are the expected node degrees defined by Equation~\eqref{eq:degree}.  
     The transition probability matrix $P^s(\A)$ 
    is generally dense and encodes the connectivity information of the graph.
     An important difference to the adjacency matrix is that whereas adding or removing an edge changes only one adjacency, it can and often does result in a substantive change in many transition matrix elements. The impact on the transition probabilities thus measures the importance of an edge in the graph structure. 
    To compare transition probabilities, we use an  {\em expected likelihood kernel} \cite{DBLP:conf/colt/JebaraK03} 
     \begin{equation}
    \sum_{\nodem=1}^{\numnodes} \sum_{\noden=1}^{\numnodes} P^s(\prob{\A})_{ij} P^s(\prob{\A}')_{ij}.
     \end{equation}
  The computational cost of computing the transition probability matrix exactly is $O(n^3)$. Sampling the matrix entries can reduce the computational cost~\cite{DBLP:conf/colt/JebaraK03}. 
 \paragraph{Kernel Weights.}
 The kernel weights $\set{\lambda}$ were set by a grid search to maximize the kernel \elbo for the DGLFRM decoder. We used the same weights for Graphite and FC. Our experiments show that these weights achieve good results without tuning them specifically for these two methods. The detailed values are given in the Appendix.

\section{Graph Neural Network Comparison Methods} 
We examine the effect of kernel regularization on several GVAE architectures. Our methodology is to keep the architecture the same as published and change only the objective function for training. As the architectures are not new, we describe them briefly and refer the reader to the appendix for details. 

\paragraph{Edge-Parallel Methods} 

For the encoder as in previous work, all our GVAE methods utilize a multi-layer Graph convolutional Network (GCN) for computing the parameters of the variational posterior distribution. We assume isotropic Gaussian distributions for the prior and posterior~\cite{kipf2016variational}.

For the decoder, we evaluate the following approaches (1)\textbf{ Graphite}~\cite{graphite}, which uses a simple dot product decoder after
GCN-style iterated message passing to transform the encoder output node representations $\set{Z}$ to final node representations $\set{\Z^{*}}$. (2) \textbf{DGLFRM}~\cite{osbm_gnn}, which applies a stochastic block model decoder. We combine the SBM decoder with the Graphite technique of transforming the node representations $\set{Z}$ to another set $\set{Z^{*}}$. (3) \textbf{\baseline}, a fully connected neural network $f(\set{Z}) = \prob{\A}$ that directly maps the node  representation to a probabilistic adjacency matrix. We include the vanilla FC decoder as a baseline because it avoids confounding the effect of kernel regularization with the influence of a specific decoder architecture.
 (4) \textbf{MolGAN}~\cite{de2018molgan} is a SOTA method for that generates edges independently based on a GAN architecture.


\paragraph{Auto-Regressive Methods.}

We utilize two SOTA Auto-Regressive architectures. (1) \textbf{GRAN.} An Auto-Regressive graph generative model that
generates one block of nodes and associated edges at a step with $O(N)$ decision steps~\cite{DBLP:conf/nips/LiaoLSWHDUZ19}.
(2) \textbf{GraphRNN.} An Auto-Regressive framework that generates the adjacency matrix sequentially. Each step generates one entry or one column through an RNN with $O(N^2)$  decision steps. In all experiments we utilized Multivariate Bernoulli version of the GraphRNN~\cite{you2018graphrnn}.

All models are trained with backpropagation for 3,000 epochs, which is similar to previous work~\cite{you2018graphrnn,DBLP:conf/nips/LiaoLSWHDUZ19} and sufficient for convergence in our datasets. For other hyperparameter settings we used those given in the original papers. The appendix contains details on the settings. We used the original papers' public repository; hence no consent was needed to curate this study.
\section{Training/Test Data and Metrics Reported}
\label{hyper3}
\begin{figure}[h]
\centering
\includegraphics[width=0.4\textwidth]{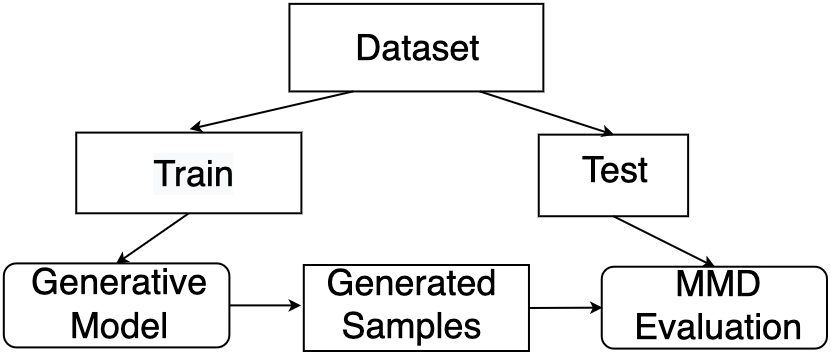}
\caption{Each dataset is split into train and test sets. The train set is then given to the generative model, and the trained model generates new sample graphs. The set of samples is then compared, to the test set using with respect to global graph features. \label{fig:test}}
\end{figure}

 Our design closely follows previous experiments on generating realistic graph structures, especially~\cite{you2018graphrnn,DBLP:conf/nips/LiaoLSWHDUZ19}, in the datasets and evaluation metrics used. 
 None of the datasets used for this research study contain any personally identifiable information or offensive content.
Following~\cite{you2018graphrnn,DBLP:conf/nips/LiaoLSWHDUZ19}, we utilize two sets of synthetic graphs and one set of real graphs.  
\begin{description}
\item[Grid] Includes 100 synthetic 2D graphs with $100\leq|\V|<400$. 
\item[Lobster]  Includes 100 synthetic trees with $10\leq|\V|\leq100$. Generated using the code from~\cite{you2018graphrnn}.
\item[Protein] Consist of  918 real-world protein graphs with $100\leq|\V|\leq500$~\cite{dobson2003distinguishing}.
\end{description}
{\em Training/Test Split.} We follow~\cite{you2018graphrnn} and use their code from a public repository to generate synthetic graphs, and randomly split the graphs sets into training graphs (80\%),  and $T$ testing graphs (20\%), see Figure~\ref{fig:test}.  We used the same train and  test graph sets for all models. After training a graph VAE model, we generate $T$ new graphs from the model~\cite{kipf2016variational,DBLP:conf/nips/MaCX18} independently of the training graphs.

{\em Evaluation Metrics.} 
  We measure how close the generated graphs are to the test graphs with respect to a set of structural graph features.  For each feature, this distance is measured by Maximum Mean Discrepancy (MMD) 
  using a Gaussian kernel exactly as in \cite{you2018graphrnn}. 
    We use the same three structure metrics as \citet{you2018graphrnn} and add graph sparsity. 
    
 \begin{description}
 \item[Degree distribution:] The probability distribution of node degrees over the graph. This metric compares the the hard adjacency matrices of test graphs and generated graphs.
    \item[Clustering coefficient distributions:] The probability distribution of the number of closed triplets\footnote{A triplet consists of three nodes that are connected by either two (open triplet) or three(closed triplet) undirected ties.} to the total number of triplets.
     \item[Average orbit counts:] The number of occurrences of all orbits (motifs) with 4 nodes.
     \item [Sparsity:] The number of zero-valued elements divided by the total number of elements in adjacency matrix representation of the  graph.
 \end{description}
\section{Experimental Results}\label{GPU}
We evaluate the quality of generated graph structures, by visual inspection and by structure metrics. We also compare graph generation speed. 
\subsection{Quality of Generated Structures}
\begin{figure}
     \centering
     \begin{subfigure}[b]{.16\textwidth}
         \centering
         \includegraphics[width=\textwidth]{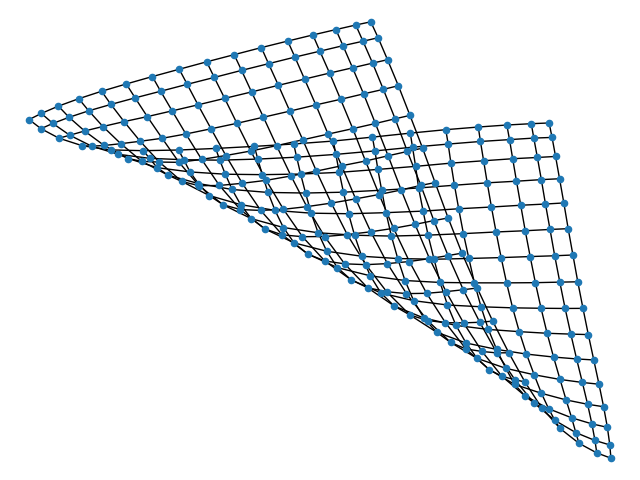}
     \end{subfigure}
          \hfill
          \begin{subfigure}[b]{.15\textwidth}
         \centering
         \includegraphics[width=\textwidth]{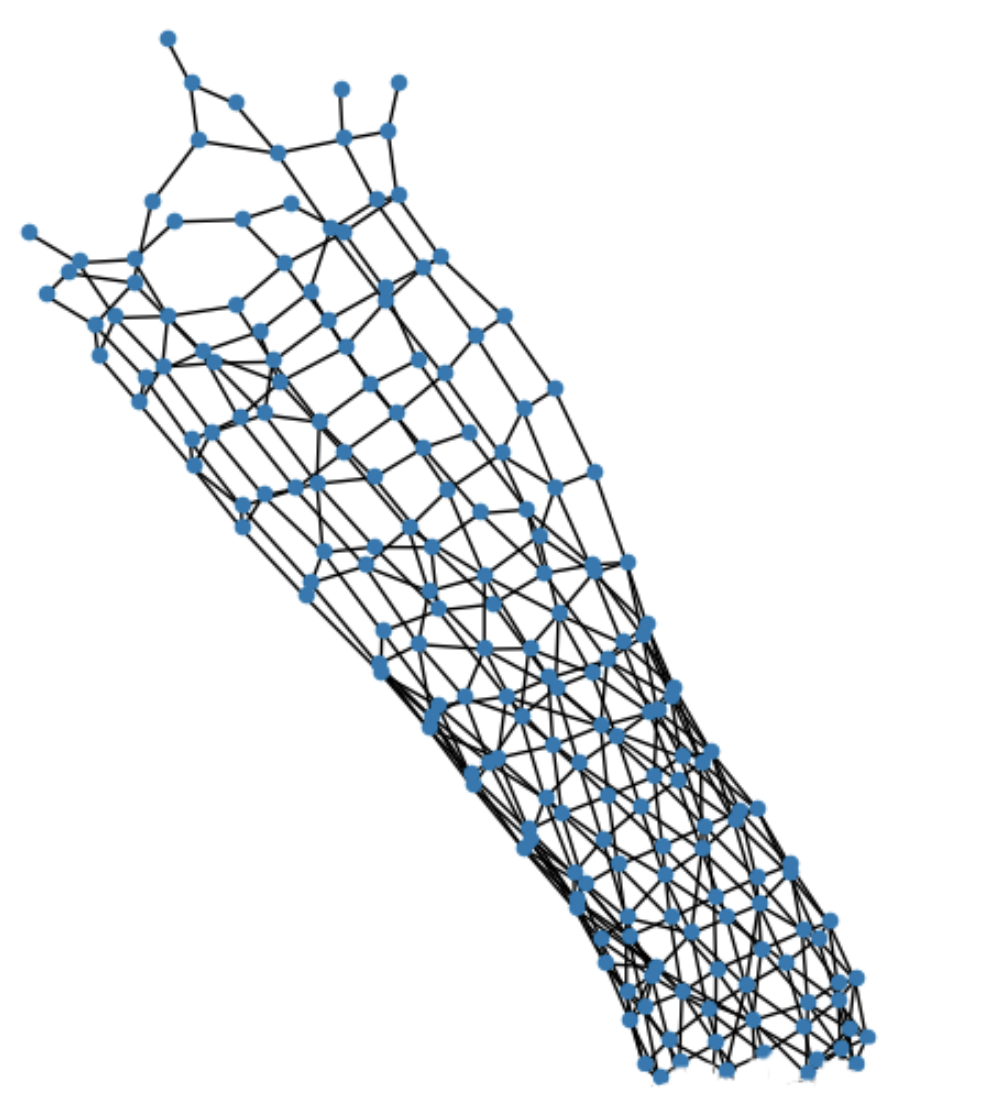}
     \end{subfigure}
          \hfill
          \begin{subfigure}[b]{.15\textwidth}
         \centering
         \includegraphics[width=\textwidth]{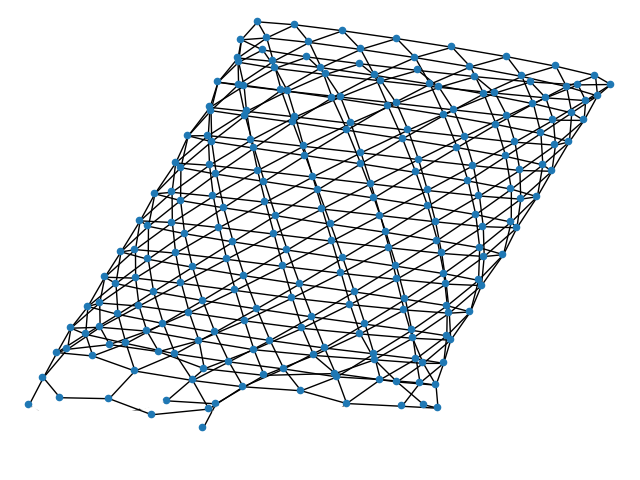}
     \end{subfigure}
               \hfill
     \begin{subfigure}[b]{0.15\textwidth}
         \centering
         \includegraphics[width=\textwidth]{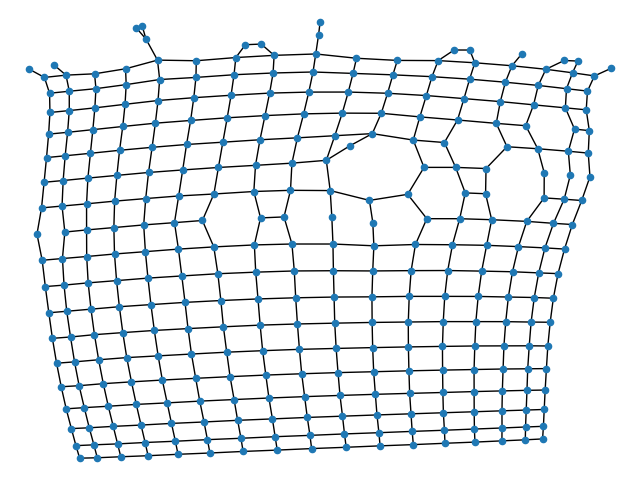}
     \end{subfigure}
          \hfill
     \begin{subfigure}[b]{0.15\textwidth}
         \centering
         \includegraphics[width=\textwidth]{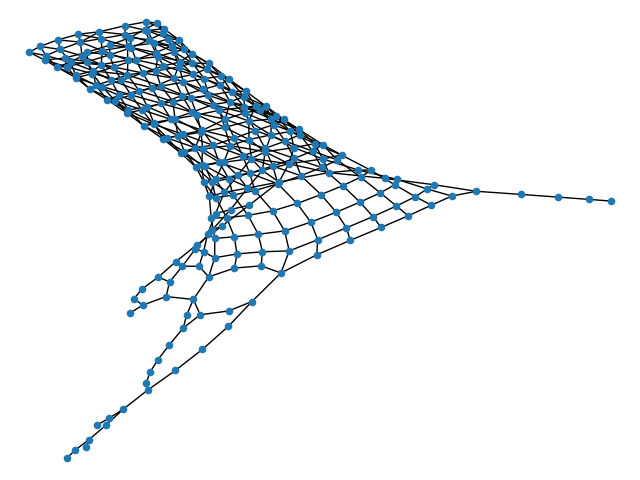}
     \end{subfigure}
     \hfill
     \begin{subfigure}[b]{0.19\textwidth}
         \centering
         \includegraphics[width=\textwidth]{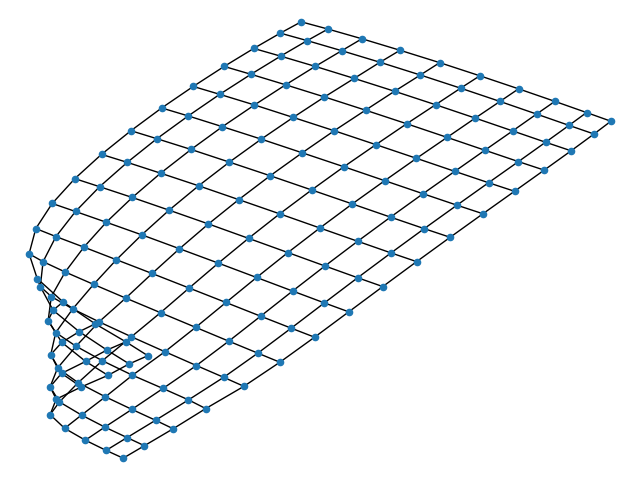}
     \end{subfigure}
         \begin{subfigure}[b]{.16\textwidth}
         \centering
         \includegraphics[width=\textwidth]{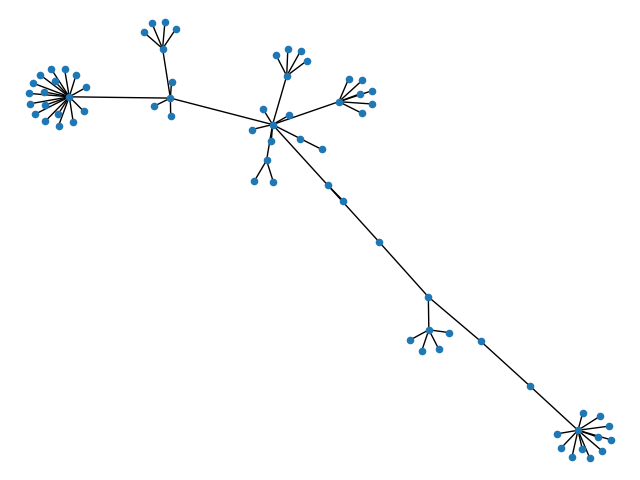}
         \caption{Test}
         \captionsetup{font=footnotesize}
         \label{fig:lobstersamples}
     \end{subfigure}
          \hfill
     \begin{subfigure}[b]{0.15\textwidth}
     \captionsetup{font={footnotesize}}
         \centering
         \includegraphics[width=\textwidth]{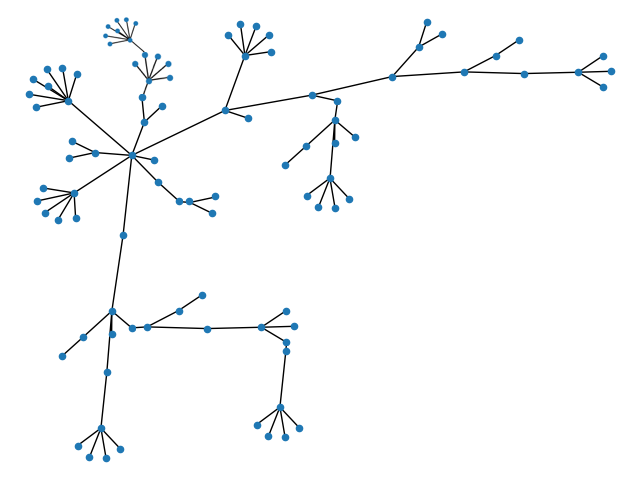}
         \caption{MolGAN}
         \label{fig:molgan_lobster}
     \end{subfigure}
          \hfill
     \begin{subfigure}[b]{0.15\textwidth}
         \centering
         \captionsetup{font={footnotesize}}
         \includegraphics[width=\textwidth]{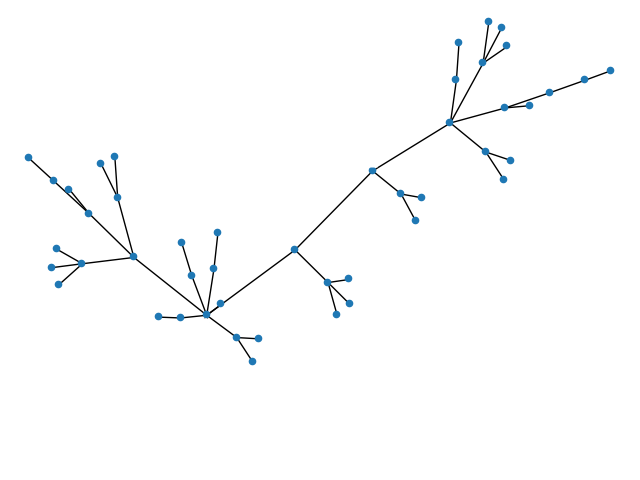}
         \caption{Graphite}
         \label{fig:graphite_lobster}
     \end{subfigure}
          \hfill
     \begin{subfigure}[b]{0.15\textwidth}
         \centering
         \captionsetup{font={footnotesize}}
         \includegraphics[width=\textwidth]{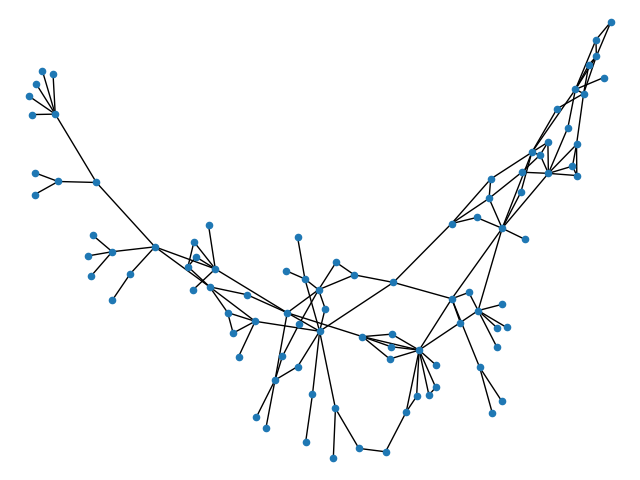}
         \caption{\baseline}
         \label{fig:lobsterFC}
     \end{subfigure}
              \hfill
     \begin{subfigure}[b]{0.15\textwidth}
         \centering
         \captionsetup{font={footnotesize}}
         \includegraphics[width=\textwidth]{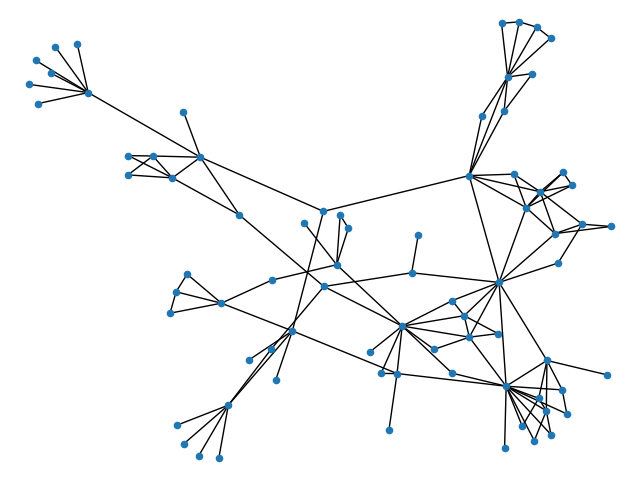}
         \caption{DGLFRM}
         \label{fig:lobster_DGLFRM}
     \end{subfigure}
     \hfill
     \begin{subfigure}[b]{0.19\textwidth}
         \centering
          \captionsetup{font={footnotesize}}
         \includegraphics[width=\textwidth]{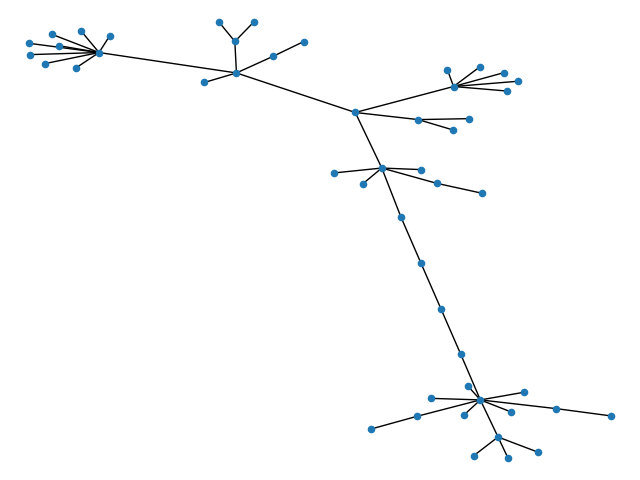}
         \caption{\small{DGLFRM-Kernel}}
         \label{fig:lobsterkernel}
     \end{subfigure} 
    \caption{Visualization of generated graphs from Grid and Lobster inputs, first and second row respectively. The left-most  column~(\ref{fig:lobstersamples}) shows 1 randomly selected target graph from the test set of each dataset.  Sample graphs were generated independently of the target (from the prior) by each method named. Each column shows the visually most similar sample graph.  DGLFRM-Kernel~(\ref{fig:lobsterkernel}) matches the target graph the best, and much better than~(\ref{fig:lobster_DGLFRM}) DGLFRM without kernel regularization.}
        \label{fig:qualitative_comparision_grid}
\end{figure}

{\em Qualitative Evaluation.} 
Figure~\ref{fig:qualitative_comparision_grid}  provides a visual comparison of randomly selected test graphs, whose structure is the same as that of the training graphs, and generated graphs. 
More examples are provided in the Appendix. 

{\em Quantitative Evaluation.}
Table \ref{table:edge-sampling} shows the MMD  for parallel-edge models (VAE and GAN) models. 
 The table shows a very large MMD improvement from kernel regularization. 
For the previous GVAE methods, {\em kernel regularization reduces MMD by at least 1 order of magnitude on almost all datasets}. The improvement for FC is less pronounced because it has by far the largest number of decoder parameters for fitting the training graphs. Generally the most improved metric is sparsity. 
As Table~\ref{table:edge-sampling2} shows, the improvement stems from generating sparser graphs that achieve realistic densities. 

Lesion studies in the appendix investigate the effects of the kernels in more detail. 1) The results of removing one kernel component at a time shows that {\em both kernel components} (degree and transition probability kernels) reinforce each other and contribute to matching the input graph gestalt. 2) Removing both kernel components shows that kernel regularization {\em improves the local reconstruction quality:} Matching global graph features 
works as a co-training objective for local link reconstruction.
\begin{table}[h]
  \caption{The impact of kernel regularization on edge-parallel models for graph generation. For the named graph structural feature, each column reports the MMD between the test set graphs (left-most column in Figure~\ref{fig:qualitative_comparision_grid}) and the generated graphs (see other columns in Figure~\ref{fig:qualitative_comparision_grid}). 
  }
  \label{table:edge-sampling}
  \centering
  \resizebox{\textwidth}{!}{
  \begin{tabular}{lcccccccccccc}
    \toprule
    \multirow{2}{3.5em}{\textbf{Method}} &  \multicolumn{4}{c}{\textbf{Grid}} &
    \multicolumn{4}{c}{\textbf{Lobster}} &
    \multicolumn{4}{c}{\textbf{Protein}} \\
    & \small{Deg.} & \small{Clus.} & \small{Orbit}& \small{Sparsity} & \small{Deg.} & \small{Clus.} & \small{Orbit}& \small{Sparsity}& \small{Deg.} & \small{Clus.} & \small{Orbit}& \small{Sparsity}\\
   \midrule
   MolGAN~\cite{de2018molgan} & $1.28$ & $1.45$ & $0.86$ & $2.01e^{-4}$ & $0.73$ & $1.50$ & $0.52$ & $3.21e^{-6}$ & $1.06$ & $0.67$ & $0.93$ & $1.65e^{-3}$\\
   Graphite~\cite{graphite}  & $1.19$ & $1.52$ & $0.88$ & $1.32e^{-7}$ & $0.85$ & $1.37$ & $0.79$ & $2.58e^{-8}$ & $1.67$ & $1.24$ & $1.09$& $1.78e^{-2}$\\
    DGLFRM~\cite{osbm_gnn}  & $1.21$ & $1.69$ & $0.95$ & $4.65e^{-8}$ & $1.17$ & $1.56$ & $0.91$ & $8.83e^{-9}$ & $0.80$ & $0.95$ & $0.80$ & $2.33e^{-6}$\\
    FC & $0.66$ & $0.91$ & $0.47$ & $1.18e^{-9}$ & $0.61$ & $1.44$ & $0.18$ &   $2.34e^{-7}$ & $0.87$ & $1.74$ & $0.81$ & $4.65^{-6}$\\
   \midrule
    Graphite-Kernel & $0.48$ & $0.81$ & $\textbf{0.02}$ & $2.45e^{-9}$ & $0.12$ & $\textbf{2e}^{\textbf{-3}}$ & $0.13$ & $\textbf{1.56e}^{\textbf{-9}}$ & $0.72$ & $0.29$ & $\textbf{2.2e}^{\textbf{-3}}$ & $0.45e^{-3}$\\
  DGLFRM-Kernel  & $\textbf{0.21}$ & $0.26$ & $0.03$ & $\textbf{3.55e}^{\textbf{-11}}$ & $0.07$ & $0.24$ & $0.08$ & $3.25e^{-9}$ & $0.65$ & $\textbf{0.08}$ & $0.74$ & $\textbf{1.73e}^{\textbf{-8}}$ \\
  FC-Kernel & $0.27$ & $\textbf{0.15}$ & $0.11 $ & $1.36e^{-9}$ & $\textbf{0.06}$ & $0.38$ & $\textbf{0.04}$ &  $5.45e^{-8}$  & $\textbf{0.32}$ & $1.32$ & $0.36$ & $3.2e^{-8}$\\
    \bottomrule
  \end{tabular}%
 } 
\end{table}

  \begin{minipage}{\textwidth}
    \begin{minipage}[b]{0.53\textwidth}
    \centering
  \begin{tabular}{lccc}
    \toprule
   Method & Grid & Lobster & Protein \\
      \midrule
   Test Set & $400.90$ & $44.45$ & $604.08$ \\
   \midrule
   MolGAN~\cite{de2018molgan} & $859.12$ & $192.46$ & $21560.03$ \\
   Graphite~\cite{graphite}  & $796.38$ & $132.85$ & $12936.78$ \\
   DGLFRM~\cite{osbm_gnn} &  $742.00$ & $152.3$ & $3693.18$\\
    FC & $674.37$ & $121.62$ & $19068.93$\\
   \midrule
   Graphite-Kernel & $578.48$ & $75.47$ & $940.35$ \\
   DGLFRM-Kernel& $344.32$ & $34.60$ & $823.51$  \\
  FC-Kernel  & $587.72$ & $47.79$ & $927.81$  \\
    \bottomrule
  \end{tabular}
      \captionof{table}{Comparison to SOTA one-shot deep generative models for average number of edges in generated graphs . The first row reports the average edge number of edges in the test set. The real-world Protein graphs show the largest improvement in density matching from kernel regularization.}
  \label{table:edge-sampling2}
    \end{minipage}
      \hfill
  \begin{minipage}[b]{.43\textwidth}
    \centering
    \includegraphics[width=\textwidth]{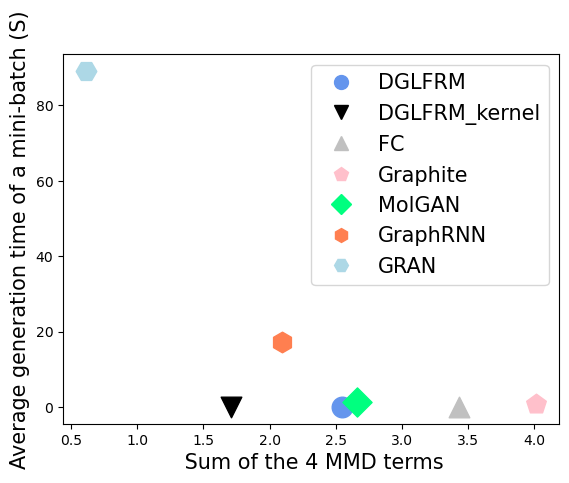}
    \captionof{figure}{Comparison of Auto-Regressive and  edge-parallel models; Generation time for a 20 graph mini-batch vs graph quality on protein data set; (0,0) is optimum.}
    \label{fig:performance_for_DD}
  \end{minipage}
  \end{minipage}

{\em Kernel Graph VAE vs. Auto-Regressive Graph Generation.} Compared to the complex Auto-Regressive methods, kernel regularization achieves competitive structure performance, with no clear pattern of superiority; the Appendix provides detailed MMD metrics.
The big advantage of edge-parallel methods is scalability: While the SOTA Auto-Regressive models are limited to generating graphs with at most 5K nodes, training graph VAEs with kernel regularization easily scales  up to 80K nodes. Figure~\ref{fig:performance_for_DD} visualizes the trade-off between generation time and structure quality by placing methods in a 2-D grid where the x-axis represents overall structural quality, measured by the sum of the 4 MMD terms, and the y-axis represents generation time of protein samples. The generation time for
each method is measured in seconds on a single GTX 1080Ti GPU. The same comparison for the other data sets appears in the Appendix. 

{\em Both Auto-Regressive methods require substantially more generation time than edge-parallel methods}, especially GRAN. GRAN is also a positive outlier in achieving an excellent aggregate  structure match to training graphs. On the aggregate structure metric, DGLFRM-Kernel actually outperforms GraphRNN. Comparing DGLFRM-Kernel to the other edge-parallel methods, the overhead from kernel computation is small.
Kernel regularization therefore offers an attractive sweet spot between scalability of training and realism of graph structures.  
        

To complement graph structure metrics, we assessed how the learned node representations support  node classification as a local prediction class. As the results in the Appendix show, node classification accuracies measured on 3 further datasets 
improves for most datasets and methods.

\section{Limitations}
\label{limitation}

The main limitation is scaling kernel regularization with the graph size, because the complexity of evaluating graph kernels grows quadratically or more with domain size (see Section~\ref{sec:input-kernels} for formulas). It may be possible to apply previous work on fast graph kernel evaluation~\cite{nikolentzos2019graph} to address this limitation. Another limitation is the need to set kernel weights as hyperparameters. 
\vspace{-0.25em} 
\section{Conclusion and Future Work}
\vspace{-0.25em}
 Within the graph VAE framework, we introduced a joint observation model for local and global graph features and derived a new \elbo objective that combines both. As the objective subtracts a kernel-based graph distance term from the standard local link reconstruction probability, we refer to it as kernel regularization. An efficient encoder-decoder graph generation architecture for kernel regularization can be obtained by extending existing graph kernels. We adapted degree and multi-step transition probability kernels for soft (reconstructed) edges.
 Empirical evaluation shows that kernel regularization improves generated graph structure by orders of magnitude compared to previous edge-parallel methods. Compared to Auto-Regressive graph generation, kernel regularization achieves competitive graph quality, and therefore offers an attractive trade-off between graph generation speed and graph structure quality. 

Kernel regularization opens a number of fruitful avenues for future work. i) A new application area for graph kernels~\cite{nikolentzos2019graph} is to investigate which are best suited to support neural graph generation. ii) Combining kernel regularization with other graph generative architectures. For example in an Auto-Regressive framework, after generating a graph an MMD term can be added to the loss function to include in backpropagation. 3) Graph embeddings~\cite{hamilton2020graph} can be used to summarize global graph features and increase the power of neural graph generators. 

In sum, modelling both global graph structure and local correlations enhances scientific modelling of graph data. In terms of learning performance, we find that they are mutually reinforcing objectives that improve each other.
\vspace{-0.25em}
\section{Societal Impact}\label{Ngimpact} 
\vspace{-0.25em}
Graph generation could have both positive and negative societal impact, depending on what the graph models. On the positive side, graphs could represent molecules, and so better modelling could aid drug discovery. On the negative side, network analysis as a field has potential to increase and misuse control over network participants. For example to motivate violations of privacy in targeting recommendations, or identify users through their social links. However, these harms can be mitigated by strengthening privacy protections during data collection. Furthermore, network analysis can provide significant societal benefit, for example by highlighting the existence and situation of marginalized communities and understanding the flow of influence and (mis)information in social networks. For a more detailed discussion about the impact of network analysis on society, see the readings by the McGill Center for Network Dynamics. The main contribution of our work supports more beneficial uses by enhancing the understanding of global network structure, as opposed to the potentially harmful targeting of individuals. 

\section*{Checklist}


\begin{enumerate}

\item For all authors...
\begin{enumerate}
  \item Do the main claims made in the abstract and introduction accurately reflect the paper's contributions and scope?
    \answerYes{}
  \item Did you describe the limitations of your work?
    \answerYes{See Section \ref{limitation} for details.}
  \item Did you discuss any potential negative societal impacts of your work?
    \answerYes{See Section \ref{Ngimpact} for details.}
  \item Have you read the ethics review guidelines and ensured that your paper conforms to them?
    \answerYes{}
\end{enumerate}

\item If you are including theoretical results...
\begin{enumerate}
  \item Did you state the full set of assumptions of all theoretical results?
    \answerYes{See Section~\ref{theory} for details.}
	\item Did you include complete proofs of all theoretical results?
    \answerYes{See Section~\ref{theory} and Appendix for details.}
\end{enumerate}

\item If you ran experiments...
\begin{enumerate}
  \item Did you include the code, data, and instructions needed to reproduce the main experimental results (either in the supplemental material or as a URL)?
    \answerYes{Code, data sets and  library requirements can be found in our repository}
  \item Did you specify all the training details (e.g., data splits, hyperparameters, how they were chosen)?
    \answerYes{See Sections~\ref{hyper3} and Appendix for all experiment settings. We also added all details, including models, generated samples, training logs and library requirements in our public repository.}
	\item Did you report error bars (e.g., with respect to the random seed after running experiments multiple times)?
    \answerNo{The reported baselines generally have training time in order of hours which make it hardly possible to run with multiple seeds on several datasets. Also the effects are large: on the key structure metrics, the proposed approach typically results in 1-2 orders of magnitude of compared metrics improvement. This evidence seems to be sufficient without error bars.}
	\item Did you include the total amount of compute and the type of resources used (e.g., type of GPUs, internal cluster, or cloud provider)?
    \answerYes{See Section~\ref{GPU} for machine details.}
\end{enumerate}

\item If you are using existing assets (e.g., code, data, models) or curating/releasing new assets...
\begin{enumerate}
  \item If your work uses existing assets, did you cite the creators?
    \answerYes{Reported data sets and codes  are  open-source and publicly available. We cited them clearly in Tables and Context}
  \item Did you mention the license of the assets?
    \answerYes{All codes and data sets used in this study  are publicly available and clearly mentioned in the Section~\ref{hyper3}.}
  \item Did you include any new assets either in the supplemental material or as a URL?
    \answerYes{We included our model implementation and  it is publicly available at -}
  \item Did you discuss whether and how consent was obtained from people whose data you're using/curating?
    \answerYes{As mentioned in  Section~\ref{hyper3}, we used the original papers' public repository; hence no consent was needed to curate this study. Also all used data sets are open source and publicly available.}
    \item Did you discuss whether the data you are using/curating contains personally identifiable information or offensive content?
    \answerYes{ As mentioned in  Section~\ref{hyper3}, None of the data sets used for this research study contain any personally identifiable information or offensive content.}
\end{enumerate}

\item If you used crowdsourcing or conducted research with human subjects...
\begin{enumerate}
  \item Did you include the full text of instructions given to participants and screenshots, if applicable?
    \answerNA{This point is not applicable for this research study.}
  \item Did you describe any potential participant risks, with links to Institutional Review Board (IRB) approvals, if applicable?
    \answerNA{This point is not applicable for this research study.}
  \item Did you include the estimated hourly wage paid to participants and the total amount spent on participant compensation?
    \answerNA{No such participants were used for this research study.}
\end{enumerate}

\end{enumerate}

\bibliography{main.bib}
\newpage
\section{Appendix} \label{Appendix}

\subsection{Proof of Proposition~\ref{prop:augmentation}.}

From the product likelihood~\eqref{eq:prod-model} we have 
\[
    p(\A|\linstance)  = \frac{1}{C(\linstance)}p(\A|\prob{\A}_{\linstance}) \cdot p(\set{\featureFunction}(\A)|\linstance))
\]
where $C(\linstance) = \sum_{\A} p(\A|\prob{\A}_{\linstance}) \cdot p(\set{\featureFunction}(\A)|\linstance))$ is a normalization constant.

We observe that 
\begin{align}
    1 &= \int_{\featureFunction} N(\featureFunction|\featureFunction(\prob{\A}_{\linstance}),\frac{1}{2 \kweight} * I) d\featureFunction \notag \\ 
    & \geq \sum_{\featureFunction:\exists \A.\featureFunction=\featureFunction(\A)} N(\featureFunction|\featureFunction(\prob{\A}_{\linstance}),\frac{1}{2 \kweight} * I) \label{eq:pdf2sum}\\
    & \geq \sum_{\featureFunction:\exists \A.\featureFunction=\featureFunction(\A)} N(\featureFunction|\featureFunction(\prob{\A}_{\linstance}),\frac{1}{2 \kweight} * I) \cdot p(\A|\prob{\A}_{\linstance}) = C(\linstance) \notag
\end{align}
The last inequality holds because $p(\A|\prob{\A}_{\linstance}) \in [0,1]$.
Hence $- \ln{C(\linstance)} \geq 0 $ and therefore
\begin{align}
 \ln{p(\set{\featureFunction}(\A)|\linstance))} - \ln{C(\linstance)} & \geq  \ln{N(\featureFunction(\A)|\featureFunction(\prob{\A}_{\linstance}),\frac{1}{2 \kweight} * I)} \notag  \\
 &= -\kweight ||\featureFunction(\A) - \featureFunction(\prob{\A}_{\linstance})||^{2} + \frac{l}{2} \ln{2 \kweight} - \frac{l}{2} \ln{2 \pi}\label{eq:normal-identity} \notag\\
 &= - \kweight \mmd_{\kernel}(\A,\prob{\A}_{\linstance})+ \frac{l}{2} \ln{2 \kweight} - \frac{l}{2}\ln{2 \pi} 
\end{align}

 Applying the general VAE lower bound to the product likelihood together with Inequality~\eqref{eq:normal-identity} yields
\begin{equation*}
\begin{split}
 \label{eq:joint-lb}
\ln{P(\A)} &\geq E_{\linstance \sim q(\Z|\X,\A)}\big[\ln{p(\A|\prob{\A}_{\linstance})} + \ln{p(\set{\featureFunction}(\A)|\linstance))} - \ln{C(\linstance)}\big] -KL\big(q_{\eparameters}(\Z|\X,\A)||p(\Z)\big) \\
&\geq E_{\linstance \sim q(\Z|\X,\A)}
\big[\ln{ p_{\dparameters}(\A|\prob{\A}_{\linstance})} - \kweight \mmd_{\kernel}(\A,\prob{\A}_{\linstance}) + \frac{l}{2} \ln{2 \kweight} - \frac{l}{2} \ln{2 \pi} \big] -KL\big(q_{\eparameters}(\Z|\X,\A)||p(\Z)\big) \\
&= \lmmd(\eparameters,\dparameters) + \frac{l}{2} \ln{2 \kweight} - \frac{l}{2} \ln{2 \pi}
\end{split}
\end{equation*}

\subsection{Graph VAEs Encoder and decoder architecture}
\label{InferenceModel}
\paragraph{Encoder}
All GVAE methods (DGLFRM, Graphite and FC) utilize a multi-layer Graph Convolutional Network (GCN) for computing the parameters of the variational posterior distribution. We assume isotropic Gaussian distributions for the prior and posterior~\cite{kipf2016variational}.
The inference model equations are as follows.
\begin{align*}
p(\set{Z}) &= \prod_{i = 1}^N {p(\set{z}_{\unode})}  = \prod_{i = 1}^N N(\set{z}_{\unode}| 0,I)\\
q({\set{Z}}\left| {\set{X},\set{A})} \right. & = 
\prod_{i = 1}^N {q(\set{z}_{\unode}| {\set{X},\set{A})}}  = \prod_{i = 1}^N {N(\set{z}_{\unode}| {\mu_i,\sigma^{2} _i *{\rm{I}})}} \\
\set{\mu},\set{\sigma} & = GCN(\A,\X)
\end{align*}
$\set{\mu}_{\unode}$ and $\set{\sigma}_{\unode}$ denote the mean and  standard deviation vectors for the node $\unode$ posterior,
\begin{equation}
    \mu,\boldsymbol{\sigma} = GCN(\A,\X)
\end{equation}
\paragraph{Decoder}
\label{GenerativeMdel}
A decoder is a deterministic function with signature $\decode:R^\ldim \times R^\ldim \rightarrow R$~\cite{hamilton2020graph}. 
Link reconstruction probabilities are computed by applying a sigmoid function to the decoder output.
\begin{equation*}
    [\prob{\A}_{\linstance}]_{\unode,\vnode} = \sigma(\decode(\set{\linstance}^{*}_{\unode},\set{\linstance}^{*}_{\vnode})) 
\end{equation*}
\citet{graphite} introduced the idea of transforming the output representations of the encoder $\set{Z}$ to a new node representation set $\set{\Z^{*}}$. 
While our GVAE comparison methods share the same GCN encoder, they differ in the decoder as follows.
\begin{description}
\item[Graphite] Uses a simple dot product decoder $\decode(\set{\linstance}^{*}_{\unode},\set{\linstance}^{*}_{\vnode}) = \set{\linstance}^{*}_{\unode}\bullet \set{\linstance}^{*}_{\vnode}$. 
Applies
GCN-style iterated message passing to transform $\set{Z}$ to $\set{\Z^{*}}$. 
\item[DGLFRM] Applies {\em a stochastic block model decoder}
$\decode(\set{\linstance}^{*}_{\unode},\set{\linstance}^{*}_{\vnode}) = \set{\linstance}^{*}_{\unode}\Lambda \set{\linstance}^{*\top}_{\vnode}$ where $\Lambda$ is a learnable parameter matrix~\cite{osbm_gnn}. We use a neural network $f(\Z)=\Z^{*}$ to transform the encoder node representations. 
\item[\baseline] A fully connected neural network $f(\set{Z}) = \prob{\A}$ that directly maps the node  representation to a probabilistic adjacency matrix.
\end{description}

\subsection{Hyperparameter details}\label{hyper1}

All models are trained with backpropagation for $3,000$ epochs, which is similar to previous work~\cite{you2018graphrnn,DBLP:conf/nips/LiaoLSWHDUZ19}.
For Graphite, we used the implementation provided by the authors~\cite{graphite}.
For DGLFRM and vanilla  FC, we use an architecture of 128-128-256 units for the 3-layer encoder. Both models are trained using the Adam optimizer~\cite{DBLP:journals/corr/KingmaW13} 
with a learning rate of $.0001$.  We also used the Beta-VAE \cite{higgins2016beta} approach with $\beta=20$ to adjust latent channel capacity for both models. 
 For DGLFRM  decoder, we use five hidden layers with  256$\numnodes$-1024-1024-1024-256$\numnodes$ units and $256 \times 256 ~\Lambda$ matrix.
 For vanilla  FC decoder we use   $256\nodeNum$-512-512-512-$\nodeNum^2$  architecture where $\nodeNum$~ is the size of maximum graph in data set. \\
 For the input Kernel weights see table \ref{table:lambdaWeights}.
  For all the other methods, we apply the hyperparameter settings suggested by the original papers. The GRAN authors suggest several settings; we use  block size and stride are both set to 1. 
\begin{table}
  \caption{The input Kernel weights $\set{\lambda}$ for each data-set  used in graph generation task. Note we used the same weight for all models.}
  \label{table:lambdaWeights}
  \centering
  \resizebox{10cm}{!}{
  \begin{tabular}{lcc}
    \toprule
   Data set & Transition Probability $(s=1\dots5)$ & In/Out Degree Dist \\
      \midrule
   Grid & $e^2$ & $e^{-4}$  \\
   Lobster & $e^2$ & $e^{-4}$  \\
   Protein  & $e^3$ & $2e^{-5}$  \\
    \bottomrule
  \end{tabular}%
 } 
\end{table}

\subsection{Qualitative evaluation of graph VAE models and kernel regularization approach in detatil} 
\begin{figure}
\begin{tabular}{m{0.09\textwidth} m{0.19\textwidth} m{0.19\textwidth} m{0.19\textwidth} m{0.19\textwidth}}
\centering
    \begin{tabular}{l}
    \centering \scriptsize{Test}
  \end{tabular}  &   \includegraphics[width=30mm]{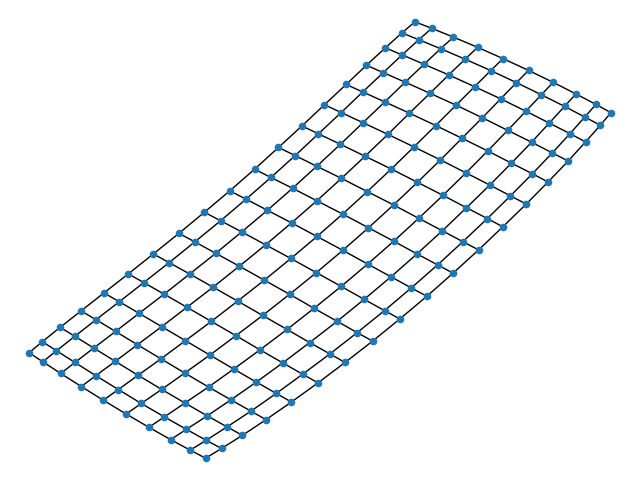} &   \includegraphics[width=30mm]{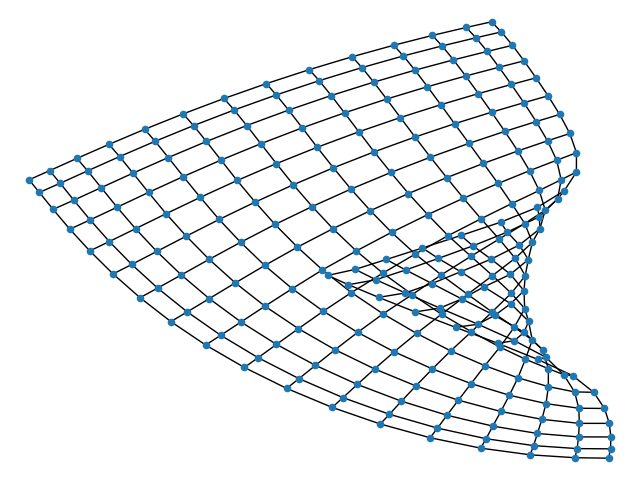}  &   \includegraphics[width=30mm]{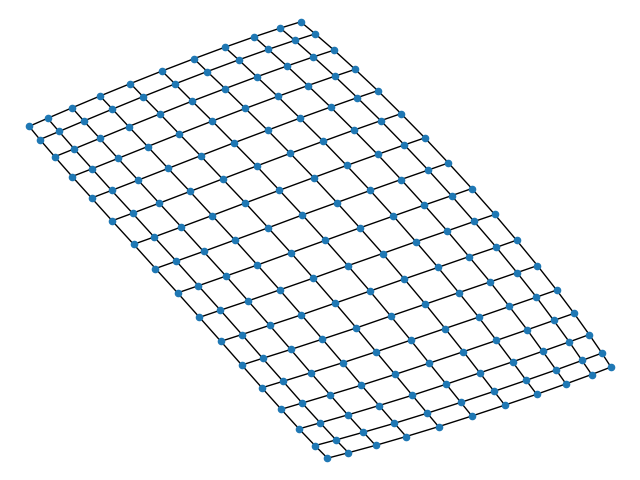} &   \includegraphics[width=30mm]{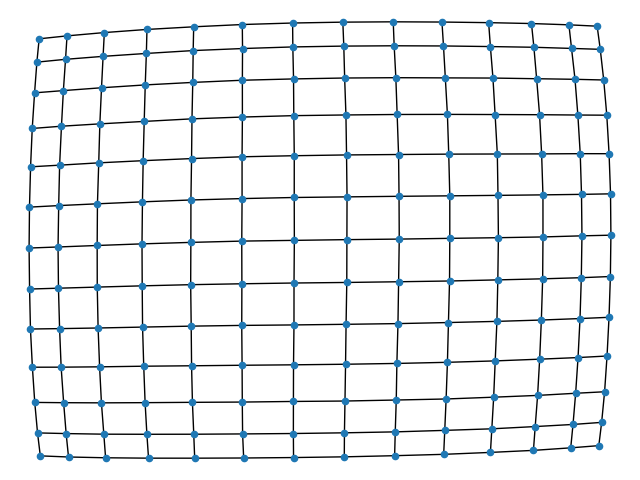}  \\ \hline
\centering
    \begin{tabular}{l}
    \centering \scriptsize{FC}
  \end{tabular}  &   \includegraphics[width=30mm]{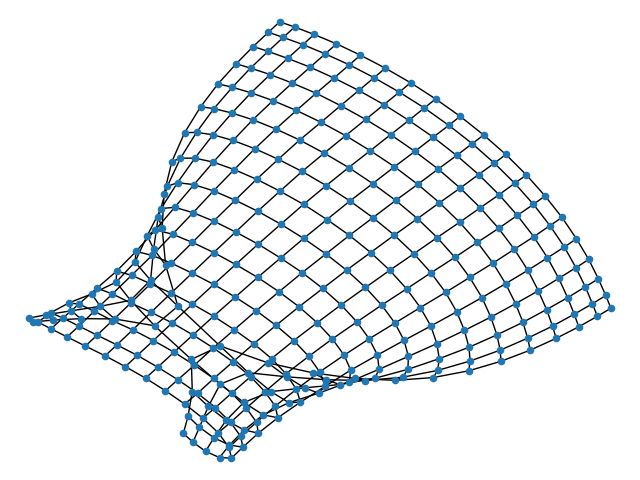} &   \includegraphics[width=30mm]{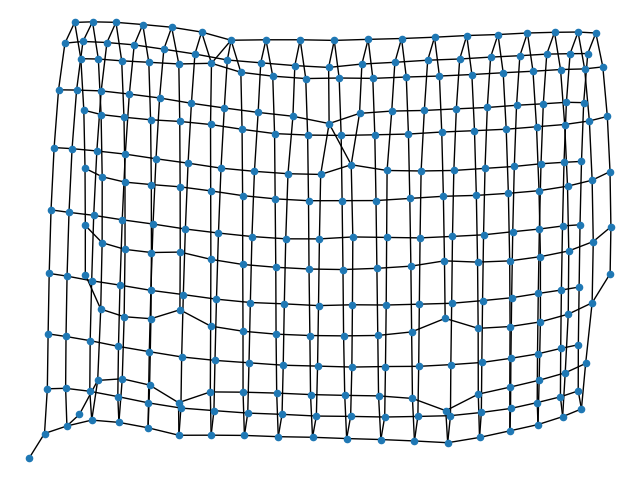}  &   \includegraphics[width=30mm]{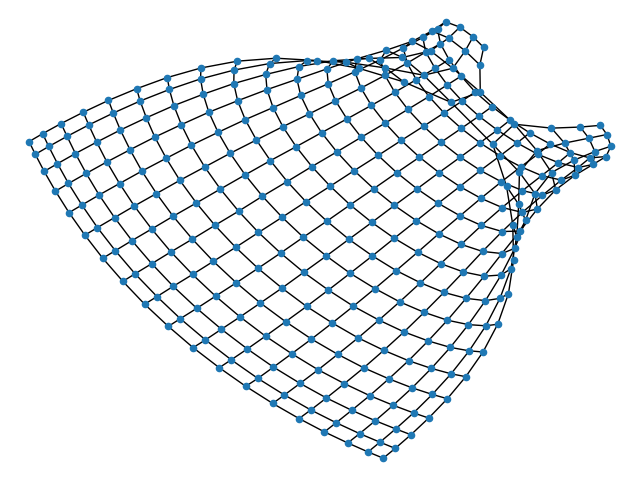} &   \includegraphics[width=30mm]{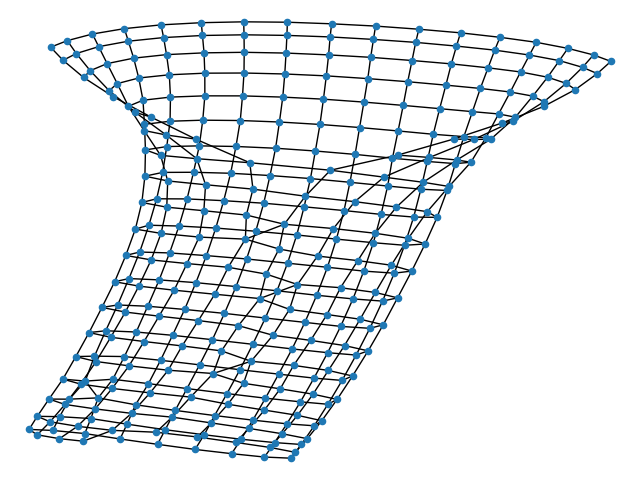}  \\
  \centering
    \begin{tabular}{l}
    \centering \scriptsize{FC-Kernel}
  \end{tabular}  &   \includegraphics[width=30mm]{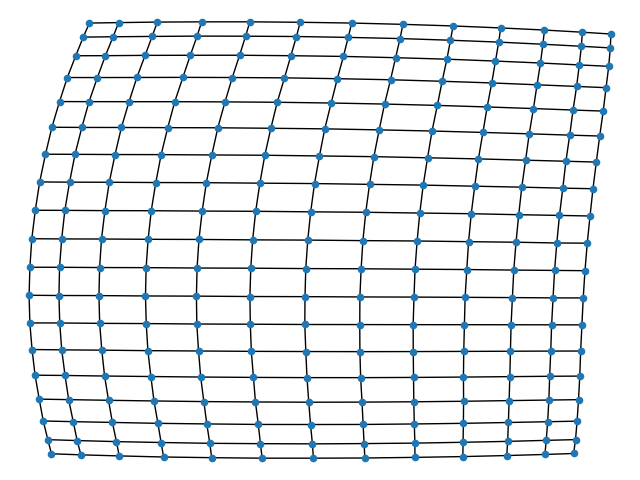} &   \includegraphics[width=30mm]{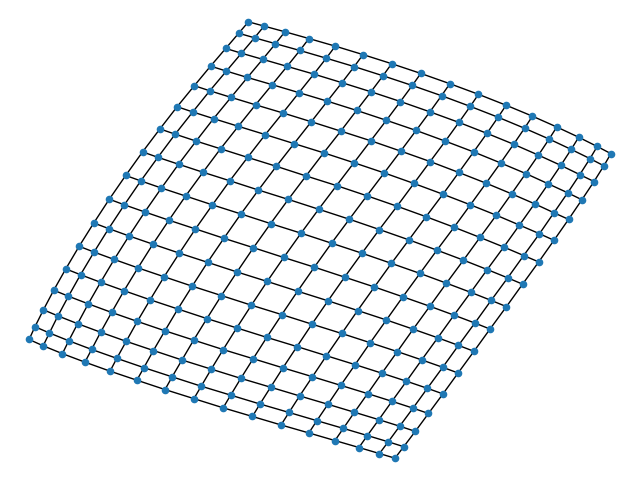}  &   \includegraphics[width=30mm]{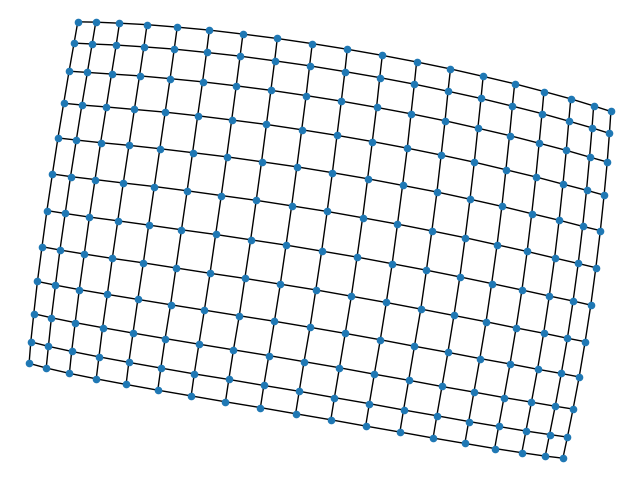} &   \includegraphics[width=30mm]{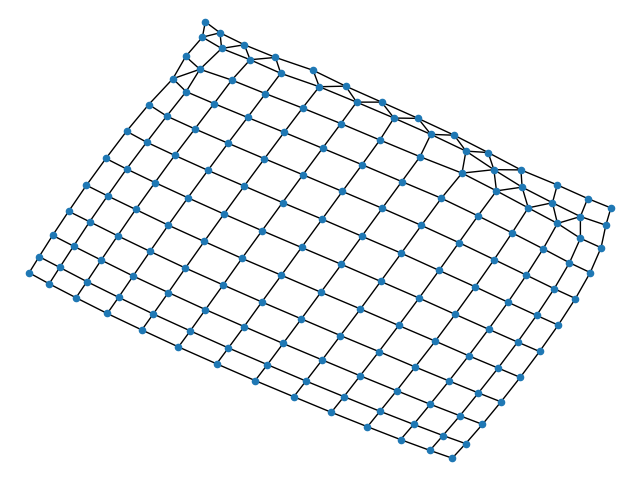}  \\ \hline
    \centering
    \begin{tabular}{l}
    \centering \scriptsize{DGLFRM}
  \end{tabular}  &   \includegraphics[width=30mm]{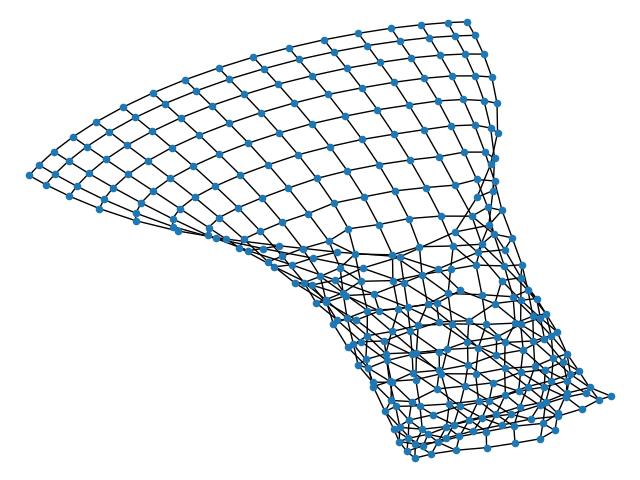} &   \includegraphics[width=30mm]{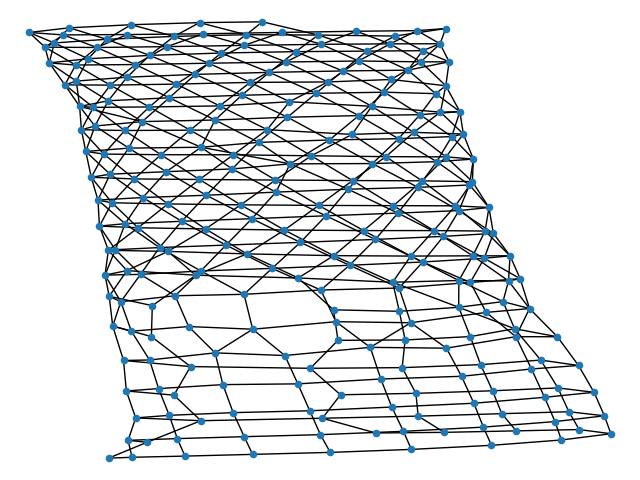}  &   \includegraphics[width=30mm]{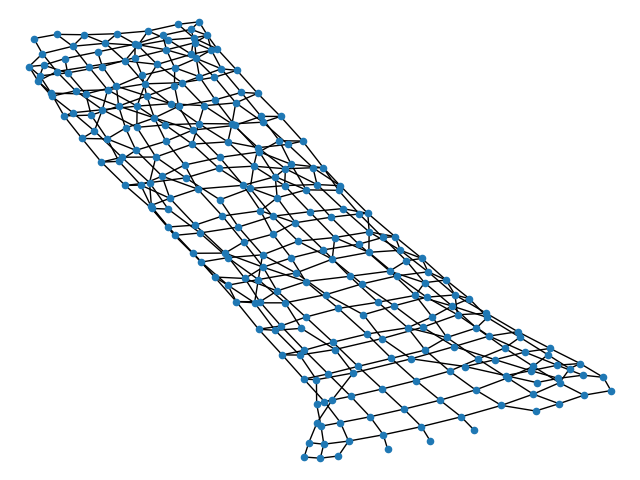} &   \includegraphics[width=30mm]{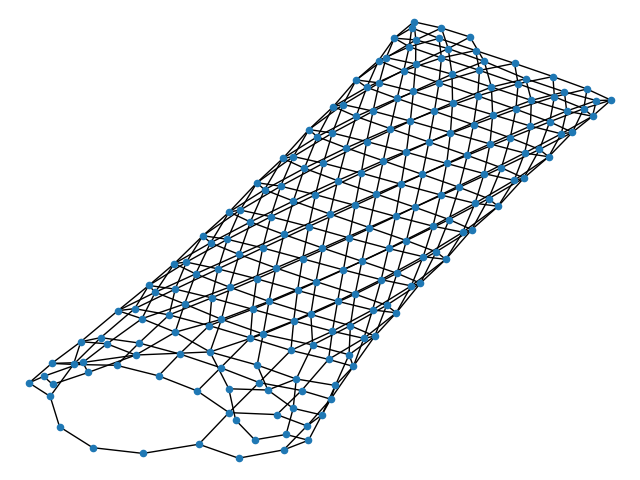}  \\
    \centering
    \begin{tabular}{l}
    \centering \pbox{15cm}{\scriptsize{DGLFRM-}\\Kernel}
  \end{tabular}  &   \includegraphics[width=30mm]{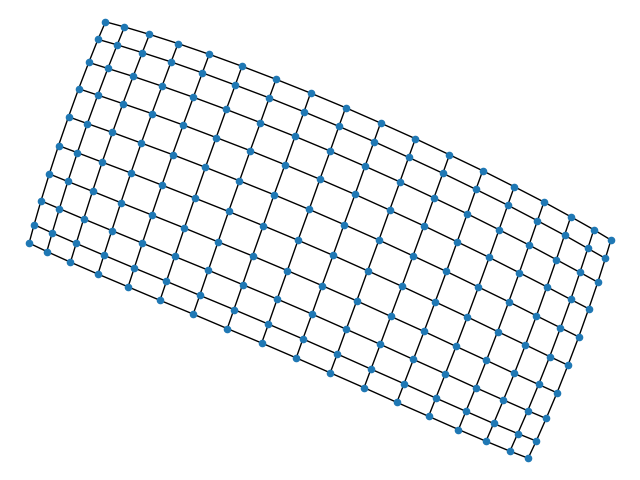} &   \includegraphics[width=30mm]{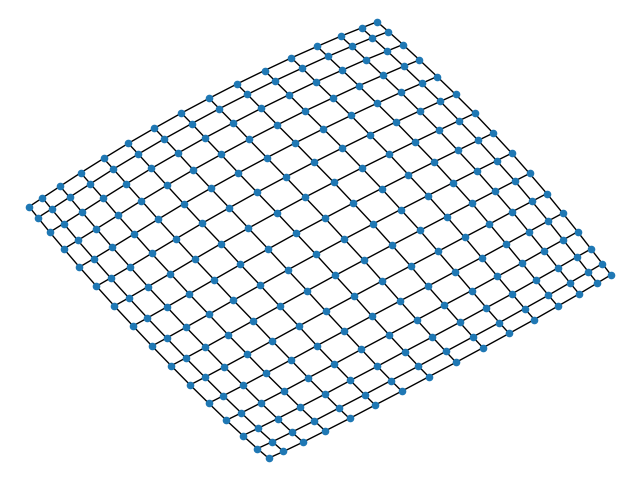}  &   \includegraphics[width=30mm]{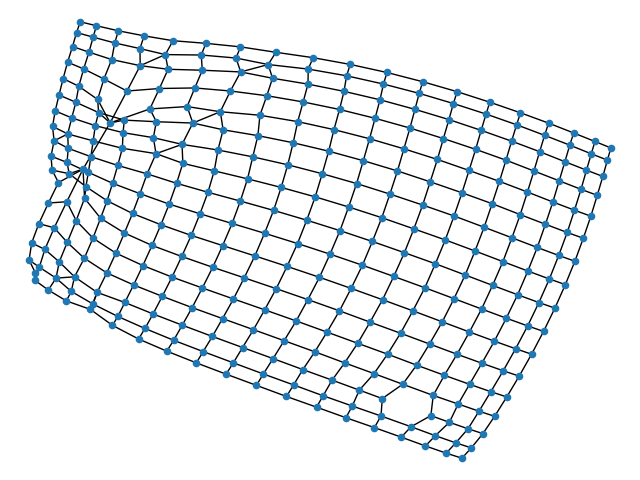} &   \includegraphics[width=30mm]{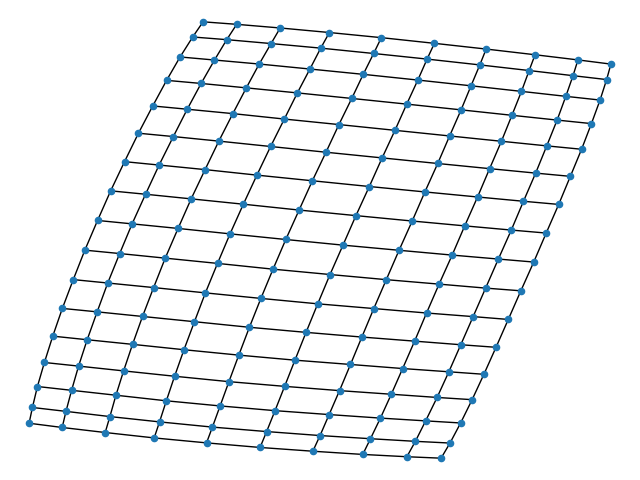}  \\
  \hline
  \centering
    \begin{tabular}{l}
    \centering \scriptsize{Graphite}
  \end{tabular}  &   \includegraphics[width=30mm]{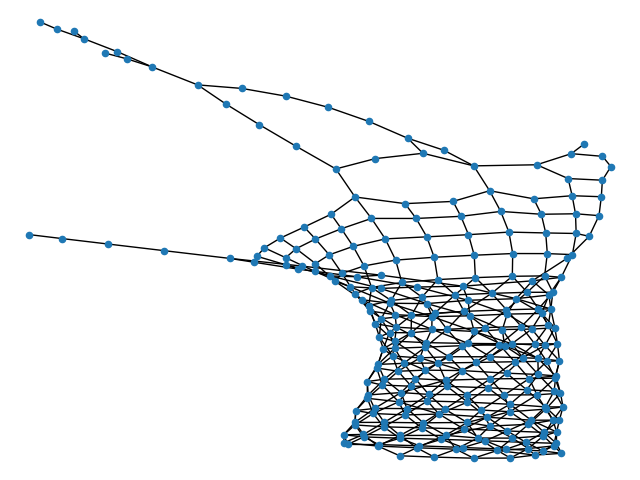} &   \includegraphics[width=30mm]{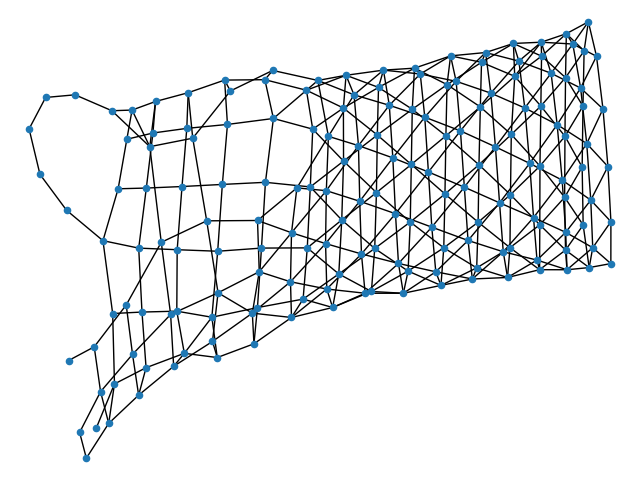}  &   \includegraphics[width=30mm]{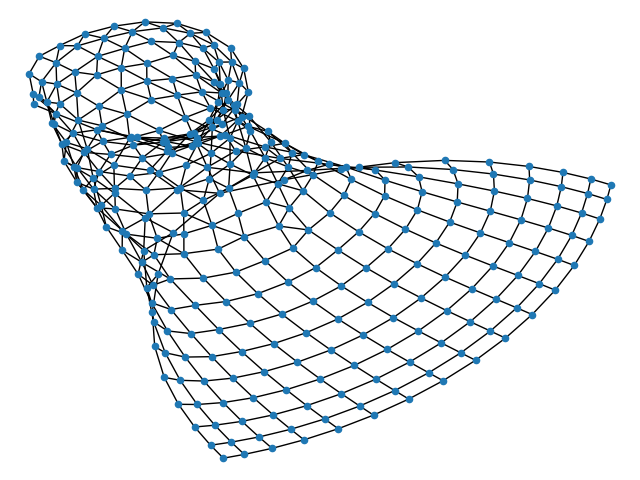} &   \includegraphics[width=30mm]{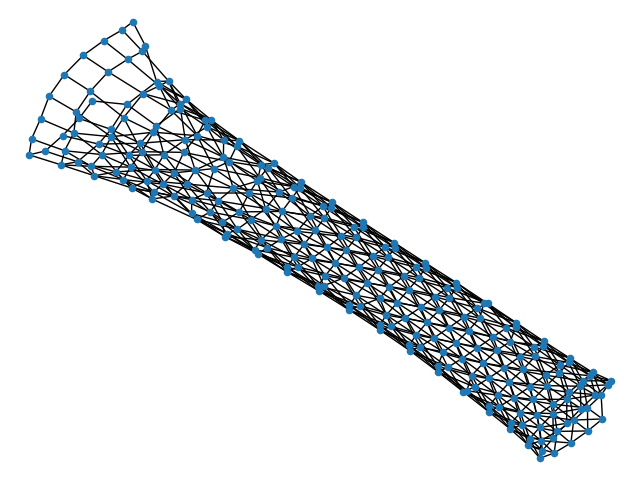}  \\
    \centering
    \begin{tabular}{l}
    \centering \pbox{15cm}{\scriptsize{Graphite-}\\Kernel}
  \end{tabular}  &   \includegraphics[width=30mm]{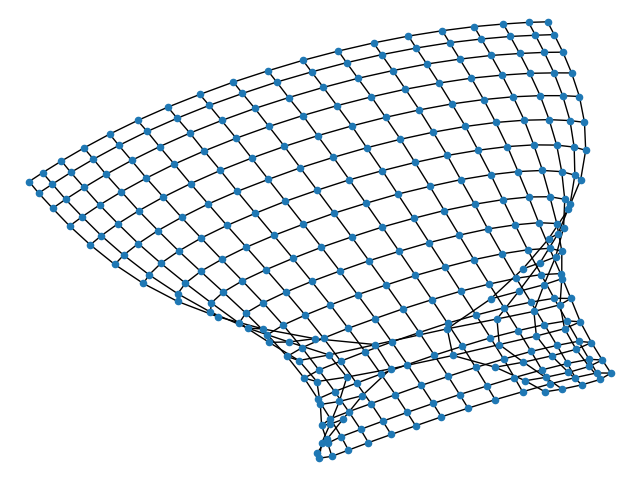} &   \includegraphics[width=30mm]{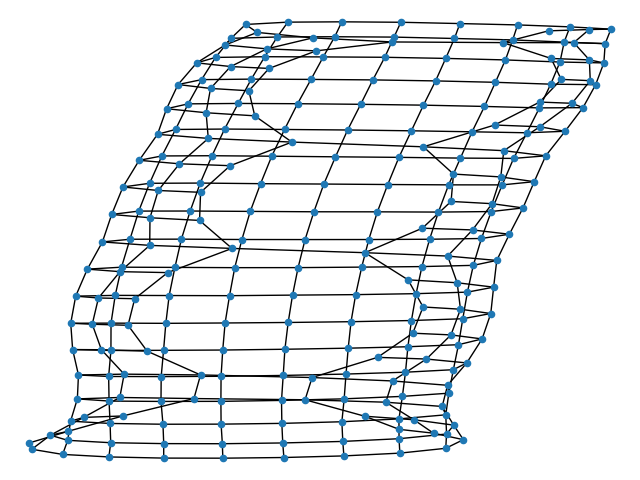}  &   \includegraphics[width=30mm]{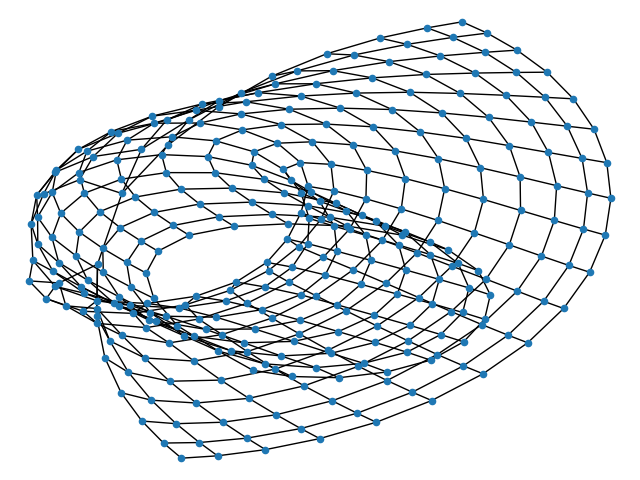} &   \includegraphics[width=30mm]{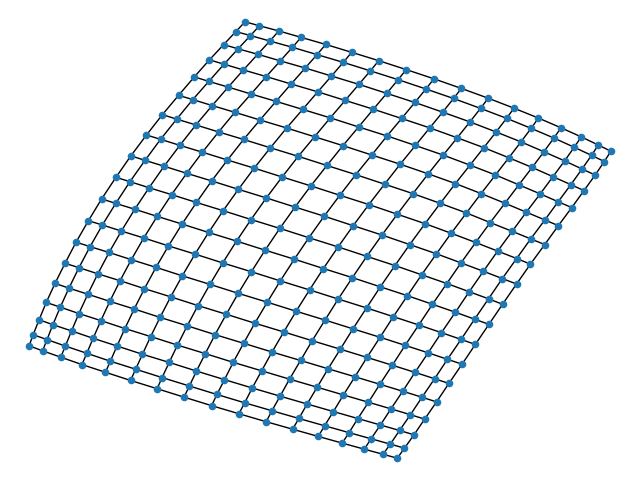}  \\
\end{tabular}
\caption{Visualization of generated graphs from Grid inputs by  SOTA graph VAE models and kernel regularization effect. The first rows show randomly selected target graphs from the test set. The first and second rows in each block show original and kernel regularized models respectively. Generated graphs by kernel regularized approach, FC-Kernel and DGLFRM-Kernel, match the target graph the best and makes a noticeable improvement in comparison to the original models. Note that generated graphs are generated independently of the target from the prior distribution $P(\linstance)$ by each method. In addition, for each model we visually select and visualize the  most similar generated samples to the test set.
}
\label{table:gridVisualization}
\end{figure}

\begin{figure}
\begin{minipage}{\textwidth}
\begin{tabular}{m{0.09\textwidth} m{0.19\textwidth} m{0.19\textwidth} m{0.19\textwidth} m{0.19\textwidth}}
\centering
    \begin{tabular}{l}
    \centering \scriptsize{Test}
  \end{tabular}  &   \includegraphics[width=30mm]{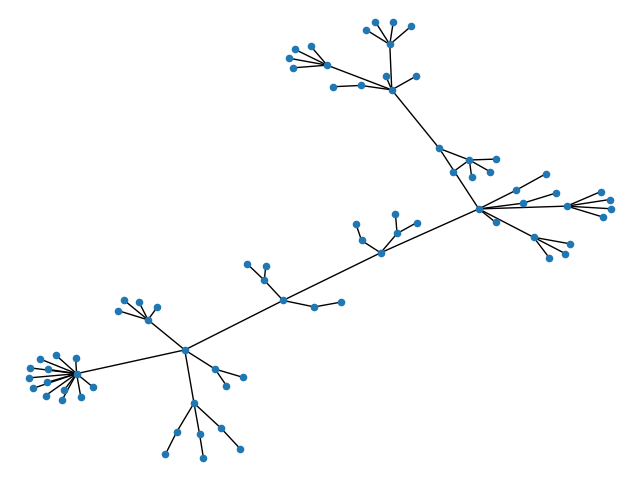} &   \includegraphics[width=30mm]{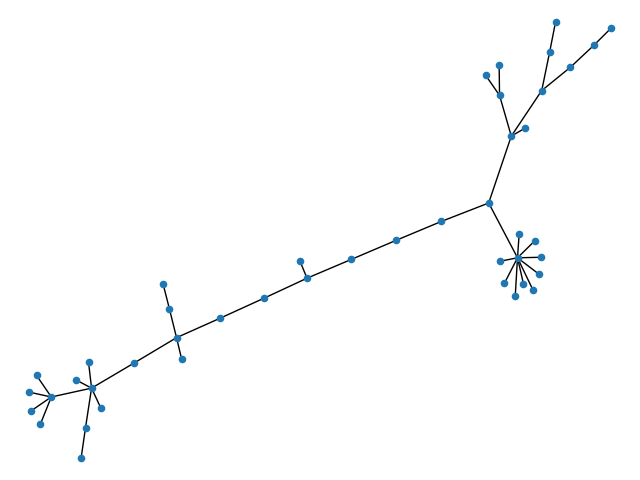}  &   \includegraphics[width=30mm]{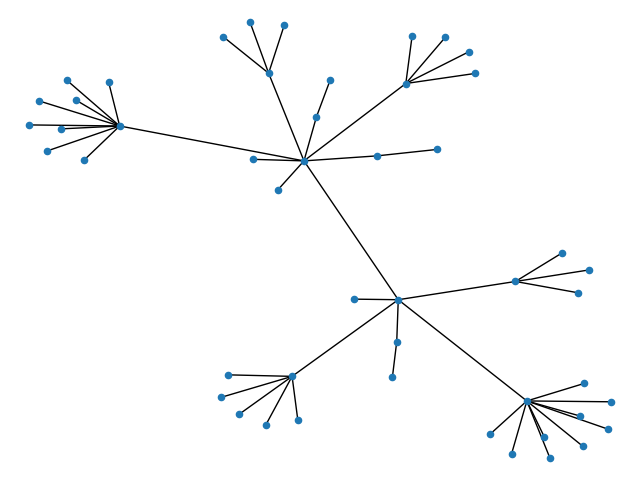} &   \includegraphics[width=30mm]{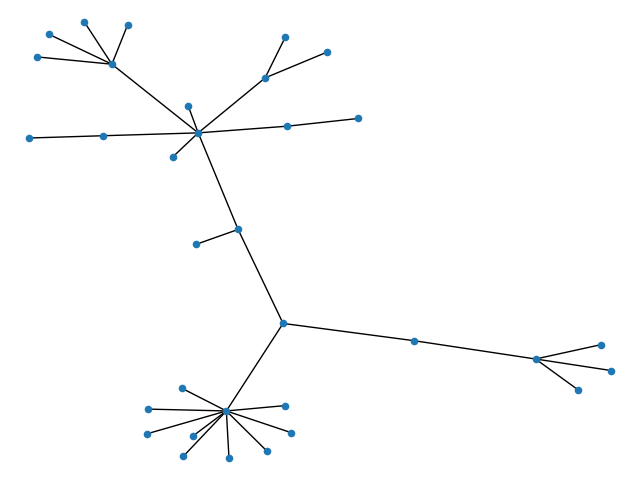}  \\ \hline
\centering
    \begin{tabular}{l}
    \centering \scriptsize{FC}
  \end{tabular}  &   \includegraphics[width=30mm]{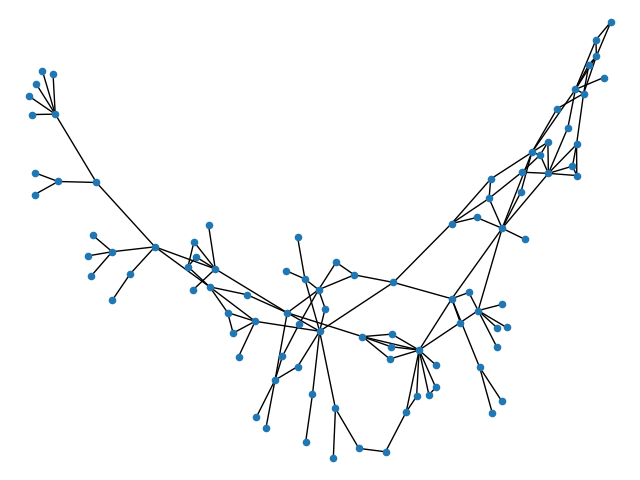} &   \includegraphics[width=30mm]{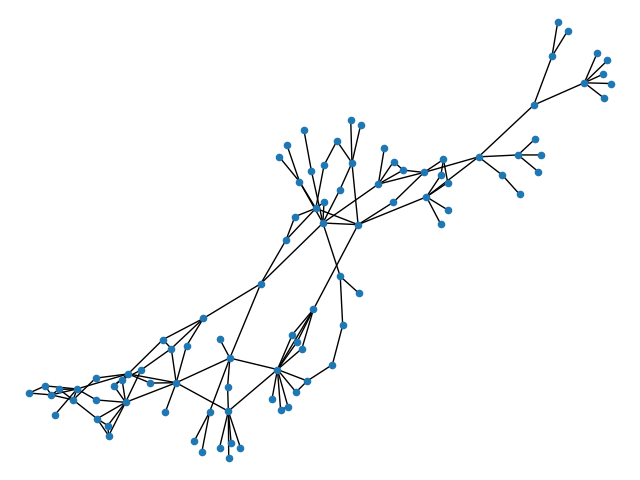}  &   \includegraphics[width=30mm]{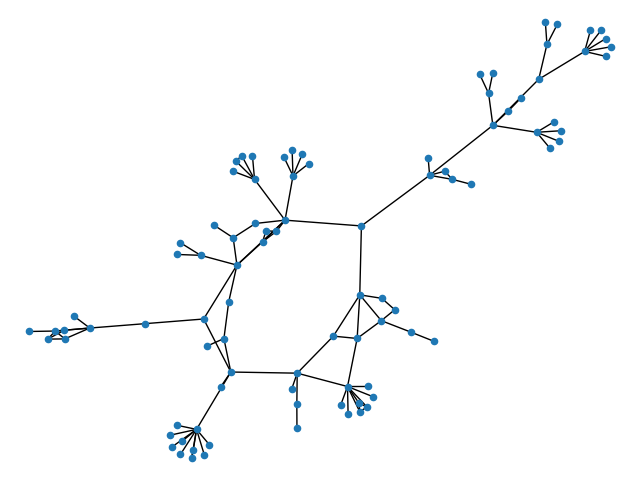} &   \includegraphics[width=30mm]{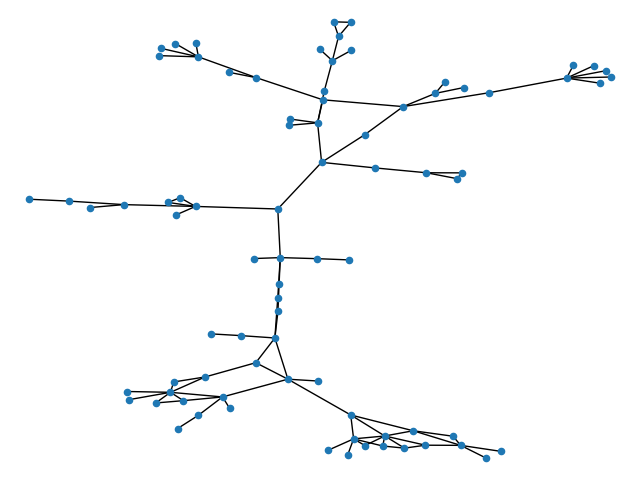}  \\
  \centering
    \begin{tabular}{l}
    \centering \scriptsize{FC-Kernel}
  \end{tabular}  &   \includegraphics[width=30mm]{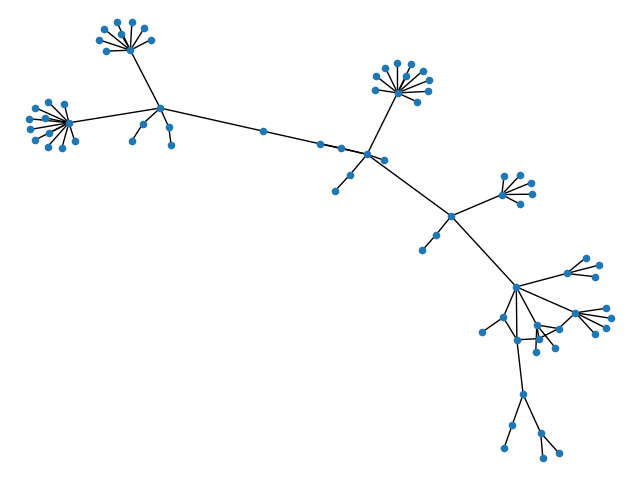} &   \includegraphics[width=30mm]{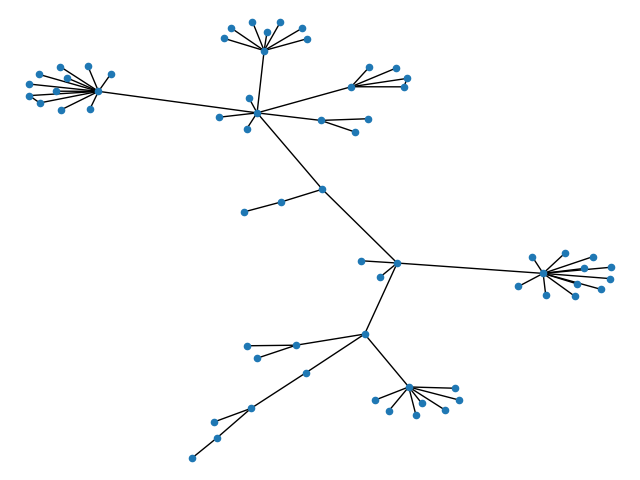}  &   \includegraphics[width=30mm]{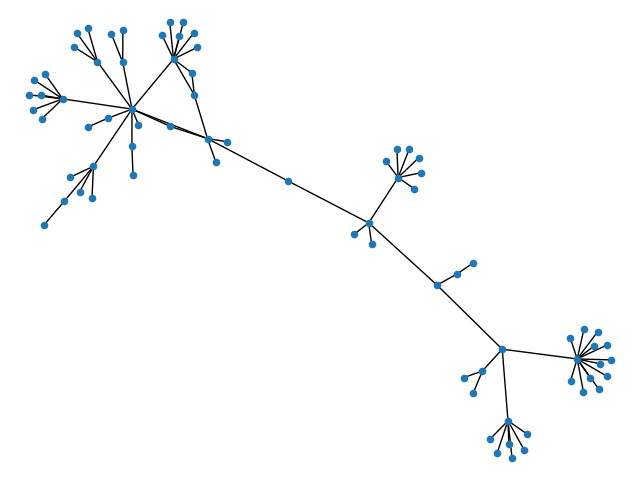} &   \includegraphics[width=30mm]{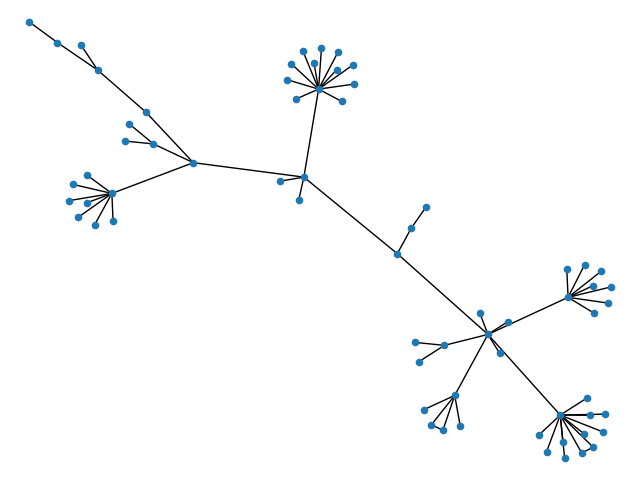}  \\ \hline
    \centering
    \begin{tabular}{l}
    \centering \scriptsize{DGLFRM}
  \end{tabular}  &   \includegraphics[width=30mm]{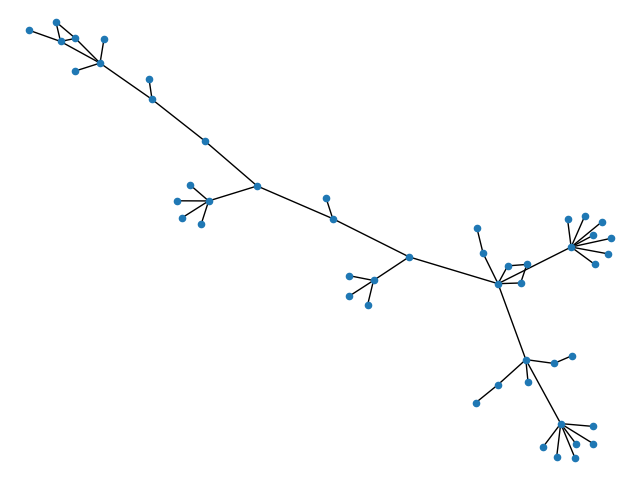} &   \includegraphics[width=30mm]{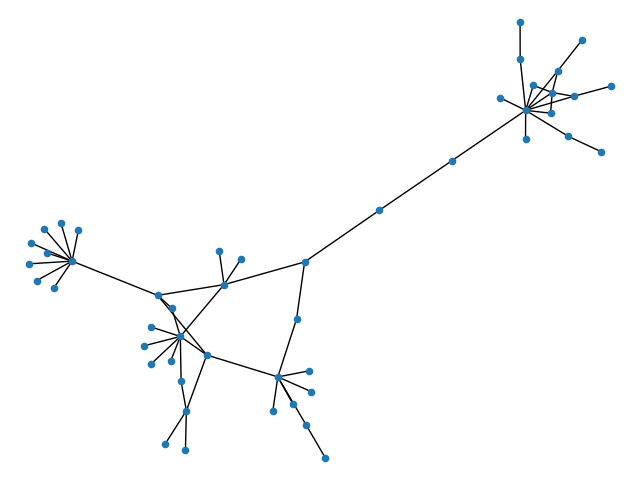}  &   \includegraphics[width=30mm]{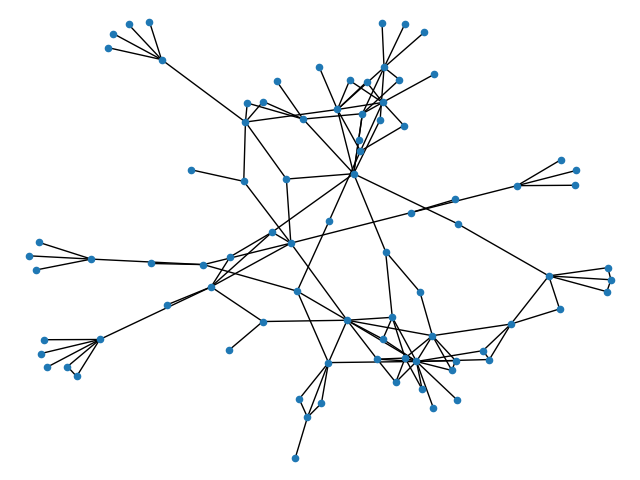} &   \includegraphics[width=30mm]{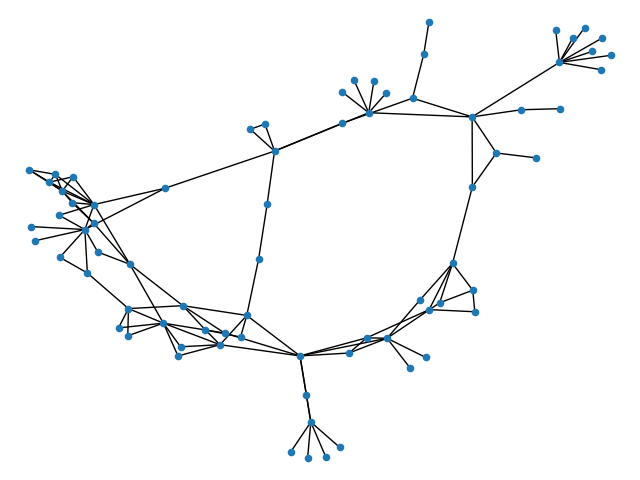}  \\
    \centering
    \begin{tabular}{l}
    \centering \pbox{15cm}{\scriptsize{DGLFRM-}\\Kernel}
  \end{tabular}  &   \includegraphics[width=30mm]{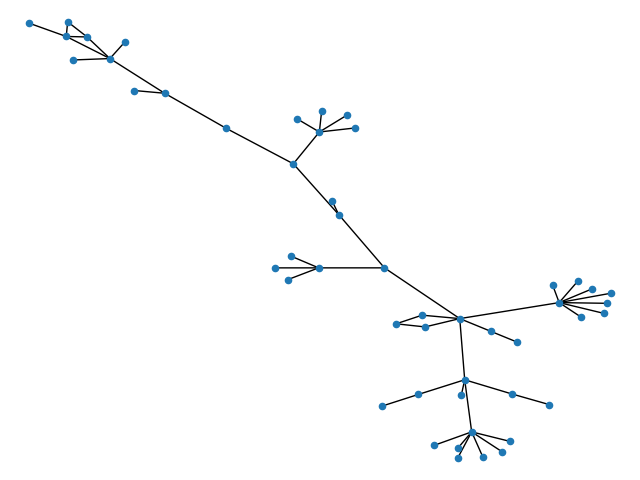} &   \includegraphics[width=30mm]{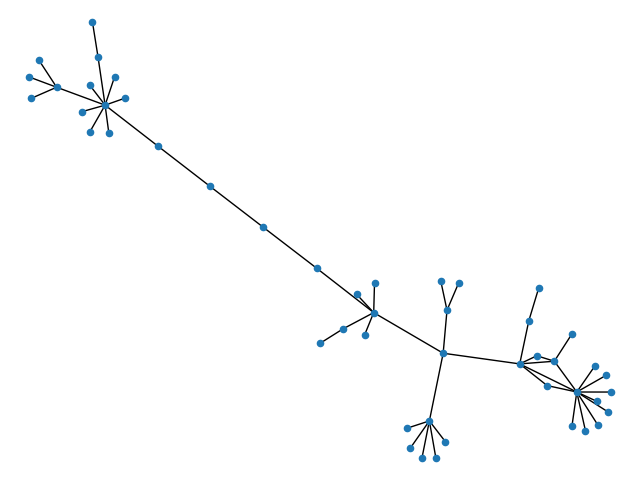}  &   \includegraphics[width=30mm]{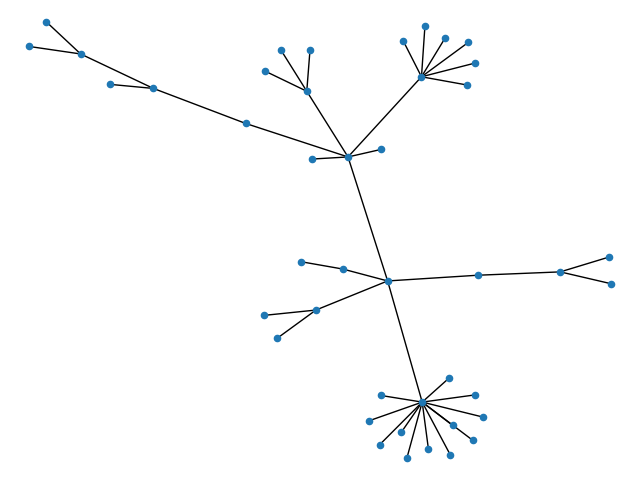} &   \includegraphics[width=30mm]{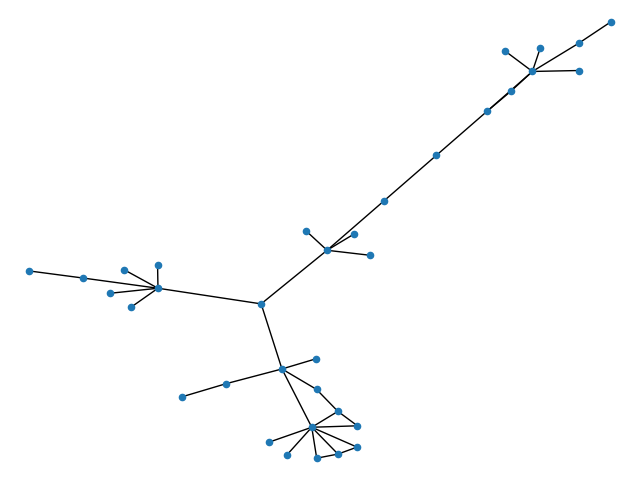}  \\ 
  \hline
  \centering
    \begin{tabular}{l}
    \centering \scriptsize{Graphite}
  \end{tabular}  &   \includegraphics[width=30mm]{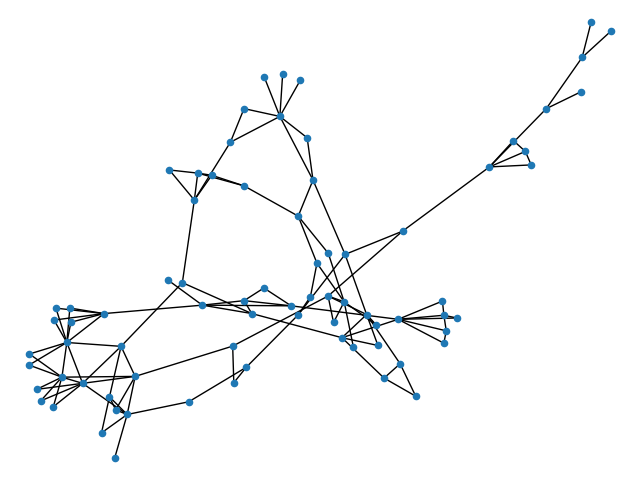} &   \includegraphics[width=30mm]{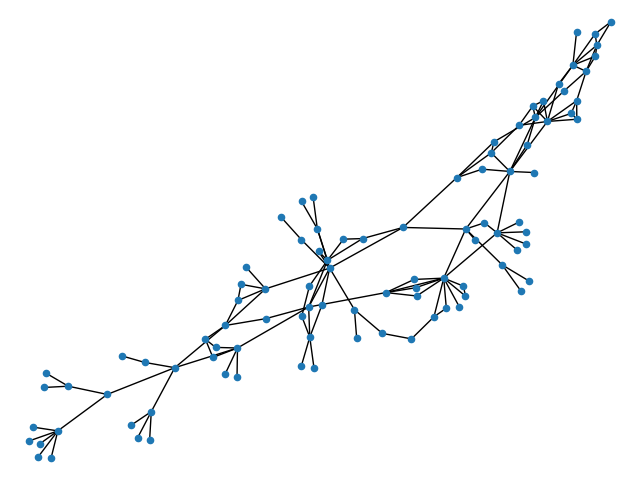}  &   \includegraphics[width=30mm]{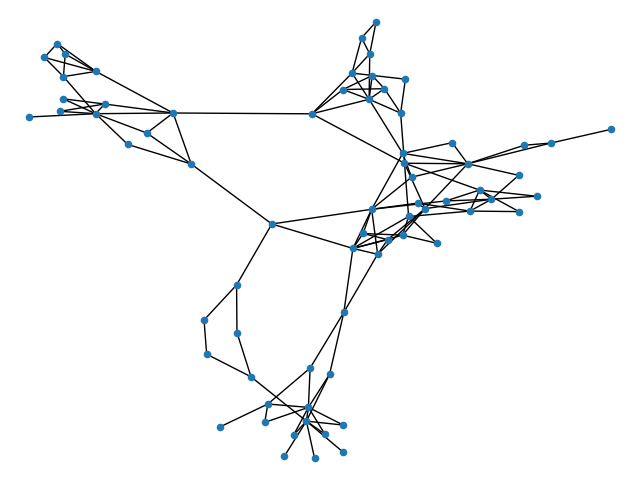} &   \includegraphics[width=30mm]{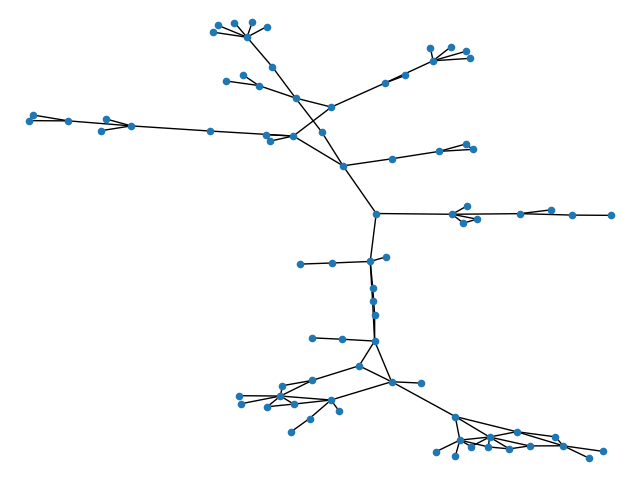}  \\
    \centering
    \begin{tabular}{l}
    \centering \pbox{15cm}{\scriptsize{Graphite-}\\Kernel}
  \end{tabular}  &   \includegraphics[width=30mm]{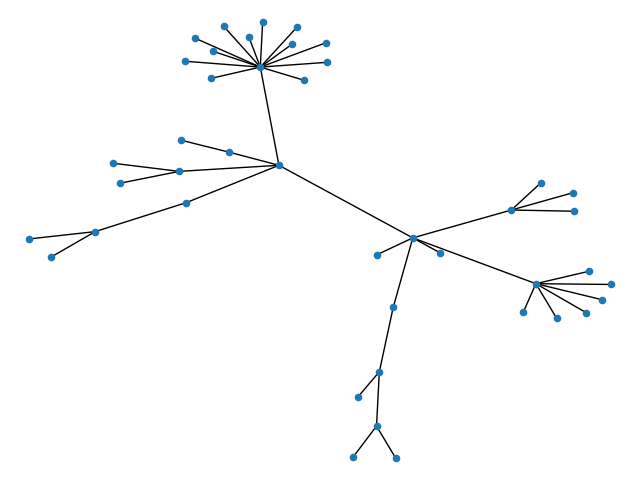} &   \includegraphics[width=30mm]{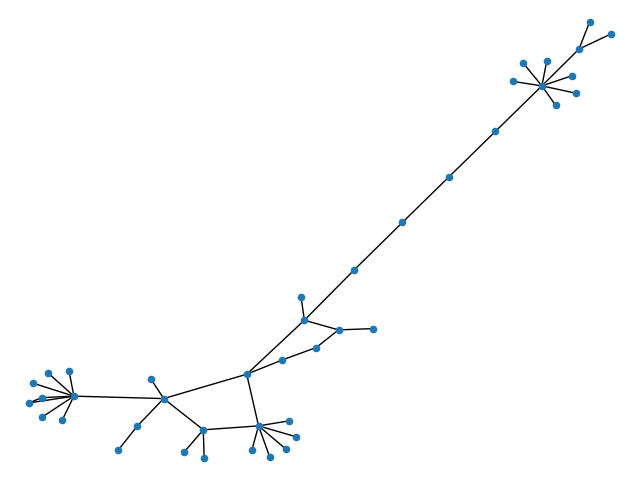}  &   \includegraphics[width=30mm]{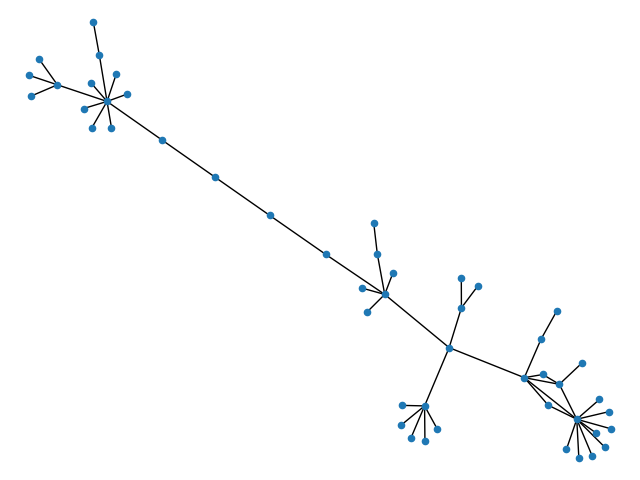} &   \includegraphics[width=30mm]{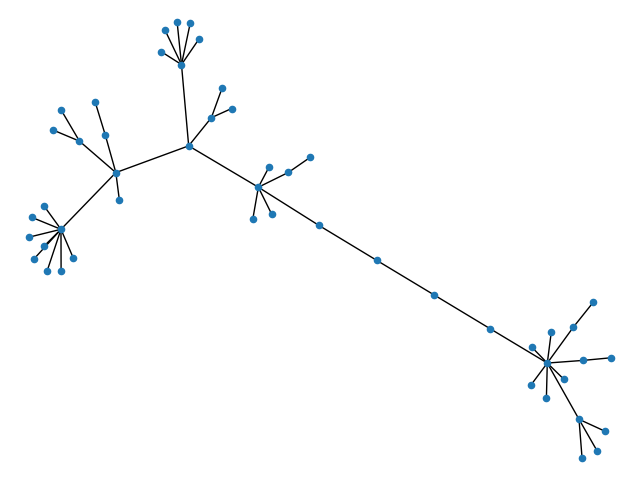}  \\
\end{tabular}
\caption[Caption for LOF]{Visualization of generated graphs from Lobster inputs, a Lobster graph is a {\em{\textbf{tree}}} in which the removal of leaves, leaves a caterpillar\footnote{A caterpillar is a graph which reduces to a path after removing its leaves.}  \cite{YAHYAEI2018}, by  SOTA graph VAE models and kernel regularization effect. The first row shows randomly selected target graphs from the test set. The first and second rows in each block show the generated graphs by original and the kernel regularized models respectively. Generated samples by kernel regularized models matched the target graph the best and  kernelized models make a noticeable improvement in comparison to the original models. Note that generated graphs are generated independently of the target, from the prior distribution $P(\linstance)$ by each method. In addition, for each model we generate $200$ samples, and visually select and visualize the  most similar ones to the test set.
}
\label{table:lobsterVisualization}
\end{minipage}
\end{figure}

We examine the effect of kernel regularization on several GVAE architectures. Our methodology is to keep the architecture the same as published and change only the objective function for training. 
Figure \ref{table:gridVisualization} and \ref{table:lobsterVisualization} provides a visual comparison of state-of-the-art GVAE architecture and the effect of kernel regularized approach. For each model we generate 200 samples from prior distribution $p(\linstance)$ and visually select and plot the most similar ones to the test set . The first row in each of the figures shows randomly selected target graphs from the test set. Each block compares the effect of the kernel regularized approach on the named model. As illustrated regularized approach noticeably improved the model's ability to generate similar graphs.

\subsection{Lesion Study on Kernel \elbo components}
In this section we drill down into the different components of the Kernel objective. Our next results study the respective importance of the input graph features used, degree distribution and transition probabilities. Table~\ref{table:lesion} shows that both kernels are necessary for achieving full structural quality. For example, compared to a single graph input feature, combining the two input graph features improves the sparsity match on grid graphs by 2 orders of magnitude, and by 1 order on Lobster graphs. Taken in isolation, matching observed transition probabilities has an even bigger impact than matching observed degree distribution---even in the Deg. columns that measure that match to degree distribution. These results provide evidence that rather than being competing objectives, matching observed graph features achieve a co-training effect where each kernel reinforces the other. 
 \begin{table}
  \caption{Lesion Study on MMD between the test set graphs and the generated graphs using DGLFRM model.}
  \label{table:lesion}
  \centering
  \resizebox{\textwidth}{!}{
  \begin{tabular}{lcccccccccccc}
    \toprule
    \multirow{2}{3.5em}{\textbf{Method}} &  \multicolumn{4}{c}{\textbf{Grid}} &
    \multicolumn{4}{c}{\textbf{Lobster}} &
    \multicolumn{4}{c}{\textbf{Protein}} \\
    & \small{Deg.} & \small{Clus.} & \small{Orbit}& \small{Sparsity} & \small{Deg.} & \small{Clus.} & \small{Orbit}& \small{Sparsity}& \small{Deg.} & \small{Clus.} & \small{Orbit}& \small{Sparsity}\\
   \midrule
   DGLFRM~\cite{osbm_gnn} & $1.21$ & $1.69$ & $0.95$ & $4.65e^{-8}$ &  $1.17$ & $1.56$ & $0.91$ & $8.83e^{-9}$& $0.80$ & $0.95$ & $0.80$ & $2.33e^{-6}$\\
   DGLFRM-Deg & $0.53$ & $0.65$ & $0.27$ & $1.13e^{-9}$ & $0.96$ & $1.43$ & $0.54$ & $4.08e^{-8}$ & $0.78$ & $\textbf{0.08}$ & $0.862$ & $1.45e^{-8}$\\
  DGLFRM-Step & $0.29$ & $0.53$ & $0.08$ & $1.66e^{-9}$ & $\textbf{0.06}$ & $0.41$ & $\textbf{0.06}$ & $8.35e^{-8} $ & $0.88$ & $1.16$ & $0.81$ & $2.15e^{-7}$\\
   DGLFRM-Kernel & $\textbf{0.21}$ & $\textbf{0.26}$ & $\textbf{0.03}$ & $\textbf{3.55e}^{\textbf{-11}}$ & $0.07$ & $\textbf{0.24}$ & $0.08$ & $\textbf{3.25e}^{\textbf{-9}}$ & $\textbf{0.65}$ & $\textbf{0.08}$ & $\textbf{0.74}$ & $\textbf{1.73e}^{\textbf{-8}}$\\
    \bottomrule
  \end{tabular}%
 }
\end{table} 
\subsection{Adjacency reconstruction vs. MMD.} 
\begin{figure}[h]
     \centering
     \begin{subfigure}[b]{.32\textwidth}
         \centering
         \includegraphics[width=\textwidth]{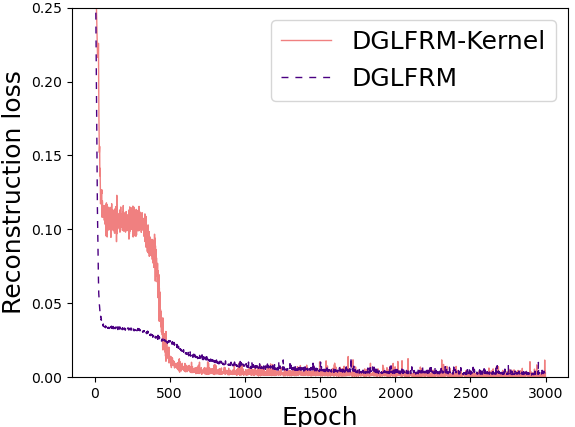}
                  \caption{Grid}
         \label{fig:performanceGrid1}
     \end{subfigure}
     \hfill
     \begin{subfigure}[b]{0.32\textwidth}
         \centering
         \includegraphics[width=\textwidth]{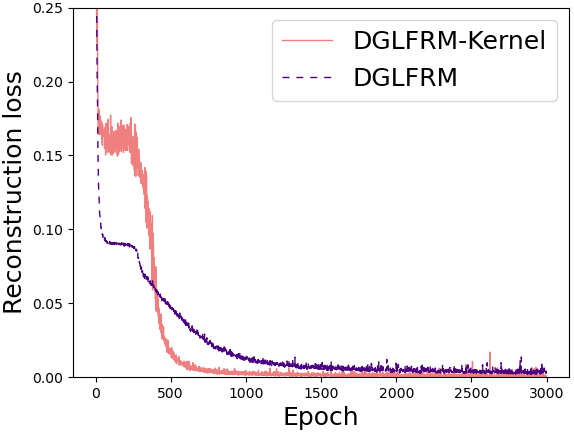}
                  \caption{Lobster}
         \label{fig:performanceLobster1}
     \end{subfigure}
     \hfill
     \begin{subfigure}[b]{0.32\textwidth}
         \centering
         \includegraphics[width=\textwidth]{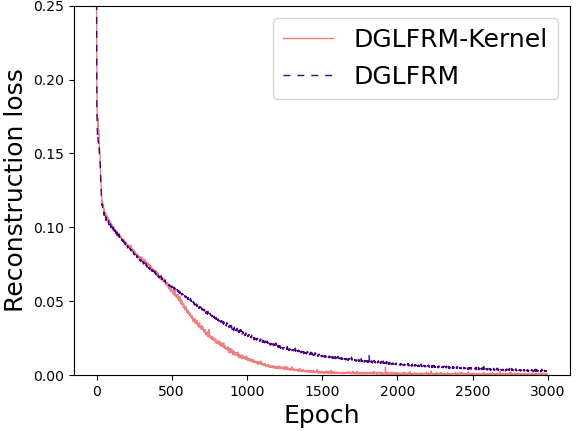}
                  \caption{Protein}
         \label{fig:performanceDD1}
     \end{subfigure}
        
     \caption{Comparing models convergence in terms of adjacency matrix reconstruction loss.}

        \label{fig:convergence}
\end{figure}

Compared to the standard graph VAE \elbo based on the cross-entropy for the {\em local} graph adjacencies, the kernel \elbo adds an MMD term to match {\em global} features. Does the MMD compete with maximizing graph adjacency reconstruction or can it be seen as a co-training objective? Comparing the standard \elbo 
\begin{align} \label{eq:old-elbo}
\lelbo(\eparameters,\dparameters) & \equiv & E_{\linstance \sim q_{\eparameters}(\Z|\X,\A)}\big[\ln p_{\dparameters}(\A|\prob{\A}_{\linstance})] -KL\big(q_{\eparameters}(\Z|\X,\A)||p(\Z)\big) 
\end{align}
with the kernelized \elbo from Equation~\eqref{eq:elbo-augment}, we find that in all datsets and for all methods, kernel regularization leads to {\em both} larger values of $\lelbo$ and $\lmmd$. This shows that the kernel loss term $\sum_{\kthFeature=1}^{\featureNUM} \set{\kweight}_{\kthFeature} \mmd_{\kernel_{\kthFeature}}(\A,\prob{\A}_{\linstance})$ serves as a co-training objective that improves the standard \elbo \eqref{eq:old-elbo} taken by itself. 
Table~\ref{table:Train_Reconstruction_comparision} gives details on the improvement in each component. We see that kernel regularization improves the local link reconstruction component and slightly increases the negative KLD. This suggests that matching global graph features helps local learning converge to a better minimum. Figure~\ref{fig:convergence} confirms this hypothesis by looking at the adjacency reconstruction loss dynamics of VAE training. We see that the adjacency reconstruction loss decreases more quickly without the MMD regularizer (FC plot). For example on grid input graphs, it falls below 0.45 already after about 100 epochs. On the other hand, convergence to an optimal reconstruction value is much more rapid with kernel regularization. For example on the grid and lobster input graphs, it occurs around 500 epochs, whereas without regularization it occurs around 1500 epochs. These results also indicate that while kernel regularization makes each step of gradient descent more costly, it reduces the overall number of steps required for convergence.
\begin{table}
  \caption{Comparing  components of the loss function. For each data set the columns show adjacency matrix reconstruction loss, $-\ln p_{\dparameters}(\A|\prob{\A}_{\linstance})$, KL divergence,  $KL\big(q_{\eparameters}(\Z|\X,\A)||p(\Z)\big)$  and kernel loss , $\sum_{\kthFeature=1}^{\featureNUM} \set{\kweight}_{\kthFeature} \mmd_{\kernel_{\kthFeature}}(\A,\prob{\A}_{\linstance})$.}
  \label{table:Train_Reconstruction_comparision}
  \centering
  \resizebox{14cm}{!}{
  \begin{tabular}{lccccccccccccccc}
    \toprule
    \multirow{2}{3.5em}{\textbf{Method}} &  \multicolumn{3}{c}{\textbf{Grid}} &
    \multicolumn{3}{c}{\textbf{Lobster}} &
    \multicolumn{3}{c}{\textbf{Protein}} \\
    & \small{Rec.} & \small{KL.} & \small{Kernel.} & \small{Rec.} & \small{KL.}  & \small{Kernel.}  & \small{Rec.}& \small{KL.} & \small{Kernel.}\\
   \midrule
   Graphite~\cite{graphite}  & $0.350$ &$1.24e^{-3}$& $2.11e^{-3}$ & $0.100$ &$2.91e^{-3}$& $1.82e^{-3}$ & $0.140$ &$0.001$& $3.12e^{-4}$ &\\
   DGLFRM~\cite{osbm_gnn}   & $0.034$ & $1.51e^{-3}$& $2.29e^{-4}$ & $0.025$& $3.50e^{-3}$ & $9.55e^{-4}$ & $0.036$& $0.002$ & $2.23e^{-4}$ &\\	
    FC  & $0.044$ & $1.56e^{-3}$ & $2.67e^{-3}$ & $0.0424$& $3.44e^{-3}$ & $3.67e^{-3}$ & $0.422$ & $0.005$& $1.04e^{-3}$ &\\
   \midrule
    Graphite-Kernel  & $0.181$ &$2.92e^{-3}$& $1.83e^{-3}$ & $0.080$ &$5.16e^{-3}$& $0.92^{-3}$ & $0.079$ &$0.007$& $2.50e^{-4}$ &\\
  DGLFRM-Kernel  & $0.015$ & $3.36e^{-3}$ & $3.14e^{-5}$& $0.007$ & $7.73e^{-3}$& $2.39e^{-5}$ & $0.002$& $0.003$ & $7.99e^{-6}$ & \\
      FC-Kernel  & $0.033$ & $5.75e^{-3}$ & $7.46e^{-5}$ & $0.0126$& $0.014$ & $3.97e^{-5}$ & $0.066$ & $0.011$ & $3.75e^{-4}$ &\\
    \bottomrule
  \end{tabular}%
 }		
\end{table}
\subsection{Comparison of Auto-Regressive methods with Kernel-Regularized graph VAE}
\begin{figure}[h]
     \centering
     \begin{subfigure}[b]{.47\textwidth}
         \centering
         \includegraphics[width=\textwidth]{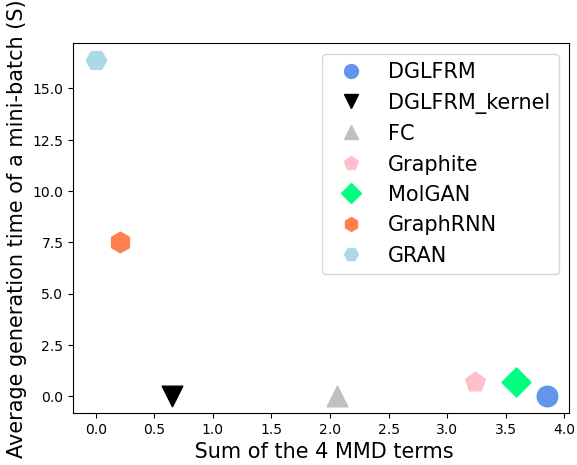}
                  \caption{Grid}
         \label{fig:performanceGrid2}
     \end{subfigure}
     \hfill
     \begin{subfigure}[b]{0.45\textwidth}
         \centering
         \includegraphics[width=\textwidth]{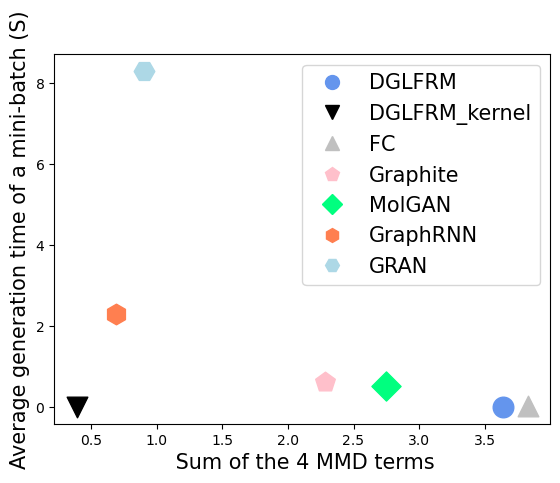}
                  \caption{Lobster}
         \label{fig:performanceLobster2}
     \end{subfigure}
     \caption{Comparison of Auto-Regressive and and edge-parallel models; The x-axis represents overall structural quality, measured by the sum of the 4 MMD terms, and the y-axis represents generation time. We use DGLFRM-kernel as Kernel regularized approach representative. While the kernel regularized model make a noticeable performance increase, has a much less generation time in comparison to Auto-Regressive models. (0,0) is optimum.}
        \label{fig:Pervormancvstraintime}
\end{figure}

Table~\ref{table:rnn} compares DGLFRM-kernel as a representative kernel-regularized method against SOTA complex deep generative graph architectures. These  methods are designed to generate realistic graph structures by generating edges sequentially, not independently, which allows for complex dependencies between new edges and edges generated so far. Given their purpose, it is not surprising that one or both of the  architectures outperform the kernel VAE on several features and dataset, as shown in Table~\ref{table:rnn}. For example,  GRAN outmatches the degree distribution of grid graphs by orders of magnitude over DGLFRM-kernel and GraphRNN. On the other hand, DGLFRM-kernel outmatches the sparsity of Lobster graphs by an order of magnitude over GRAN and GraphRNN. So we do not see a clear pattern of superiority between Auto-Regressive and kernel-regularized architectures in terms of graph structure quality. However , as mentioned earlier {\em both Auto-Regressive methods require substantially more generation time than edge-parallel methods}. Figure \ref{fig:Pervormancvstraintime} visualizes the trade-off between generation time and structure quality of Auto-Regressive and edge-parallel models on Grid and Protein data set.
 \begin{table}
 \caption{Comparison of Kernel-Regularized graph VAE with the state-of-the-art Auto-Regressive complex deep architectures models in Graph Generating task using MMD. There is no clear pattern of superiority. For all MMD metrics, the smaller the better.}
  \label{table:rnn}
  \centering
  \resizebox{\textwidth}{!}{
  \begin{tabular}{lcccccccccccc}
    \toprule
    \multirow{2}{3.5em}{\textbf{Method}} &  \multicolumn{4}{c}{\textbf{Grid}} &
    \multicolumn{4}{c}{\textbf{Lobster}} &
    \multicolumn{4}{c}{\textbf{Protein}} \\
    & \small{Deg.} & \small{Clus.} & \small{Orbit}& \small{Sparsity} & \small{Deg.} & \small{Clus.} & \small{Orbit}& \small{Sparsity}& \small{Deg.} & \small{Clus.} & \small{Orbit}& \small{Sparsity}\\
   \midrule
   GRAN~\cite{DBLP:conf/nips/LiaoLSWHDUZ19} & $\textbf{2.0e}^{\textbf{-3}}$ & $7.5e^{-3}$ & $\textbf{1.90e}^{\textbf{-5}}$ & $ 2.49e^{-10}$ & $0.20$ & $0.33$ & $0.36$ & $2.07e^{-7}$ & $\textbf{0.12}$ & $0.44$ & $\textbf{0.03}$ & $\textbf{6.59e}^{\textbf{-9}}$\\
  GraphRNN~\cite{you2018graphrnn} & $0.20$ & $\textbf{8.0e}^{\textbf{-4}}$ & $9.23e^{-3}$ & $4.77e^{-10}$ & $0.31$ & $\textbf{0.01}$ & $0.14$ & $5.01e^{-7}$ & $0.82$ & $0.88$ & $0.38$ & $1.46e^{-6}$ \\
   \midrule
  DGLFRM-Kernel  & $0.21$ & $0.26$ & $0.03$ & $\textbf{3.55e}^{\textbf{-11}}$ & $\textbf{0.07}$ & $0.24$ & $\textbf{0.08}$ & $\textbf{3.25e}^{\textbf{-9}}$ & $0.65$ & $\textbf{0.08}$ & $0.74$ & $1.73e^{-8}$ \\
    \bottomrule
  \end{tabular}%
 }
\end{table}
\subsection{Node classification}
In this setting, we train a generative graph model on a single, typically large, input graph~\cite{xiang2011relational}. While generative graph models like VAEs can be evaluated in both the multi-graph and single graph input settings~\cite{graphite,hamilton2020graph}, most studies of graph VAEs are based on a single input graph. 
 The learned node representations are typically evaluated on node classification or edge prediction tasks~\cite[Ch.1]{hamilton2020graph}, two key local prediction tasks. We conduct experiments for node classification tasks with three single graph data sets to study the effect of kernel regularization on local prediction. Our results also provide evidence that kernel regularization does not lead to overfitting, in that classifier performance on test nodes with unseen node labels improves for almost all models and datasets. Our results show a similar pattern as in the multi-graph setting.

\begin{table}[h]
  \caption{ Statistics of the single graph data sets. }
  \label{table:Datastat}
  \centering
  \begin{tabular}{cllcc}
    \toprule
     Dataset & Nodes & Edges & \#Node Type & \#Edge Type 
    \\
    \midrule\midrule
    
   IMDb & 12,772 & 19,120 & 3 & 2 
   \\
    \hline
 DBLP & 18,405 & 47,283 & 3 & 2
      \\\hline
      
   ACM & 8,993 & 18,929 & 3 & 2 
       \\
    \bottomrule
  \end{tabular}
\end{table}

\begin{table}[h]
  \caption{Evaluating the effect of kernel matching in  node classification task. See Table \ref{table:Datastat} for dataset details.}
  \label{table:nodeClassification}
  \centering
  \resizebox{12cm}{!}{
  \begin{tabular}{lcccccc}
    \toprule
    \multirow{2}{3.5em}{\textbf{Method}} &  \multicolumn{2}{c}{\textbf{DBLP}} &
    \multicolumn{2}{c}{\textbf{ACM}} &
    \multicolumn{2}{c}{\textbf{IMDb}} \\
    & \small{Macro-F1} & \small{Micro-F1} & \small{Macro-F1} & \small{Micro-F1} &\small{Macro-F1} & \small{Micro-F1} \\
   \midrule
    DGLFRM  & $46.94$ & $82.20$ &  $86.27$ & $97.90$ &$\textbf{20.93}$  & $\textbf{28.36}$\\
  DGLFRM-Kernel  & $\textbf{49.29}$ & $\textbf{84.15}$ & $\textbf{90.38}$ & $\textbf{99.36}$ &$\textbf{20.93}$  & $\textbf{28.36}$\\
  \hdashline
      Graphite  & $48.04$ & $78.59$ &  $26.32$ & $65.23$ &  $27.77$  & $47.56$\\
  Graphite-Kernel  & $\textbf{65.60}$& $\textbf{80.58}$&  $\textbf{48.80}$ & $\textbf{69.95}$  & $\textbf{44.95}$  & $\textbf{61.38}$\\
    \bottomrule
  \end{tabular}%
 }
\end{table}
\paragraph{Data sets}\label{hyper5}
The single graph data sets used in this study include IMDb, ACM, and DBLP. The details of the datasets are shown in Table~\ref{table:Datastat}. None of the datasets used for this research study contain any personally identifiable information or offensive content. DBLP, ACM, and IMDb~\cite{DBLP:conf/nips/YunJKKK19} are Heterogeneous graphs and consist of three different types of nodes and two types of undirected edges.  Given the node representation learned by models, we aim to recover the node types.
\paragraph{Experimental setup}
We train models 
for $1000$ epochs. We tuned hyperparameters of $\lmmd$ for node classification task for each dataset; $\set{\lambda}$ weight for S-Step transition probability $(s=1\dots5)$ and In/Out degree distribution are $e^{-2}$ and $e^{-13}$, respectively. They are the same for all methods. For Graphite  we apply the hyperparameter settings suggested by the original papers. For DGLFRM and DGLFRM-kernel we use  Adam optimizer with a learning rate of $2e^{-4}$.  In addition for both models we use  three-layer GCNs with 128, 128 and 16 dimensions in the encoder. 
\paragraph{Node classification.} 
Following \citet{grover2016node2vec} we use node feature representations as input to a one-vs-rest logistic regression classifier.  We split the data to train and test and report the performance in terms of both Macro-F1 and Micro-F1\cite{grover2016node2vec}.
Tables \ref{table:nodeClassification} summarize Micro-F1 and Macro-F1 evaluations on three single graph data sets. The experiment shows that the Kernel \elbo objective function results in a noticeable improvement on Macro-F1 and Micro-F1 on all data sets and models except for the DFLFRM on IMDb which  is almost the same.
\subsection{Code overview}
The implementation is provided at \textit{"-"}\footnote{We will use our GitHub repository upon paper acceptance.}. main.py includes the training pipeline and also kernel Elbo and Elbo objective functions implementation. Graph kernels implementation can be found at graph-kernel.py. models.py contains the graph VAE with different  Encoder and Decoder architecture implementations. Source codes for loading real graph datasets and generating synthetic graphs are included in data.py. All the Python packages used in our experiments are provided in environment.yml.
Generated graph samples for each model and dataset are provided in the "result/" directory, both in the pickle and png format.
We also added modified implementations of the Graphite and MolGAN baselines  in the "baselines/" directory. Note that we only made a few changes in the original public repositories to be able to report the effect of the new objective function.

\end{document}